\documentclass{article}
\usepackage[preprint,nonatbib]{neurips_2024}
\usepackage[utf8]{inputenc} 
\usepackage[T1]{fontenc}    
\usepackage{algorithm}
\usepackage[noEnd=false,indLines=true]{algpseudocodex}
\usepackage{hyperref}       
\usepackage{url}            
\usepackage{booktabs}       
\usepackage{amsfonts}       
\usepackage{nicefrac}       
\usepackage{microtype}      
\usepackage{xcolor}         
\usepackage{bm}
\usepackage{tablefootnote}
\usepackage{hyperref}
\usepackage{url}
\usepackage{graphicx}
\usepackage{booktabs}
\usepackage{multirow}
\usepackage{makecell}
\usepackage{xspace}
\usepackage{bbm}
\usepackage{colortbl}
\definecolor{mycolor_blue}{HTML}{E7EFFA}
\definecolor{mycolor_green}{HTML}{E6F8E0}
\definecolor{mycolor_gray}{HTML}{ECECEC}
\definecolor{pearDark}{HTML}{2980B9}
\usepackage{subfig,graphicx}

\usepackage{amsmath,amsfonts,bm}

\def\eqref#1{equation~\ref{#1}}

\def\1{\bm{1}}

\DeclareMathAlphabet{\mathsfit}{\encodingdefault}{\sfdefault}{m}{sl}
\SetMathAlphabet{\mathsfit}{bold}{\encodingdefault}{\sfdefault}{bx}{n}

\vspace{-2pt}
\title{Infinity$\infty$: Scaling Bitwise AutoRegressive Modeling for High-Resolution Image Synthesis}
\def\methodNAME{Infinity\xspace}

\vspace{-5pt}
\author{
  \vspace{-25pt}\\
  \textbf{Jian Han\thanks{Equal contribution. $^{\dag}$Corresponding author: \href{mailto:yuanzehuan@bytedance.com}{\color{black}{yuanzehuan@bytedance.com}}}$~~$,\quad Jinlai Liu$^{*}$,\quad Yi Jiang$^{*}$,\quad Bin Yan}\vspace{3pt} \\
  \textbf{Yuqi Zhang,\quad Zehuan Yuan$^{\dag}$,\quad Bingyue Peng,\quad Xiaobing Liu}\vspace{3pt} \\
  ByteDance\vspace{3pt} \\
  \texttt{\small \{hanjian.thu123,liujinlai.licio,jiangyi.enjoy,bin.yan\}@bytedance.com,}\\ 
  \texttt{\small \{zhangyuqi.hi,yuanzehuan,bingyue.peng,will.liu\}@bytedance.com,}\vspace{8pt}  \\
  Codes and models:~\, \url{https://github.com/FoundationVision/Infinity}
  \vspace{-4pt} \\
}

\usepackage{amsthm}

\begin{document}

\maketitle


\begin{figure}[ht]
\vspace{-16pt}
\begin{center}
    \includegraphics[width=0.97\linewidth]{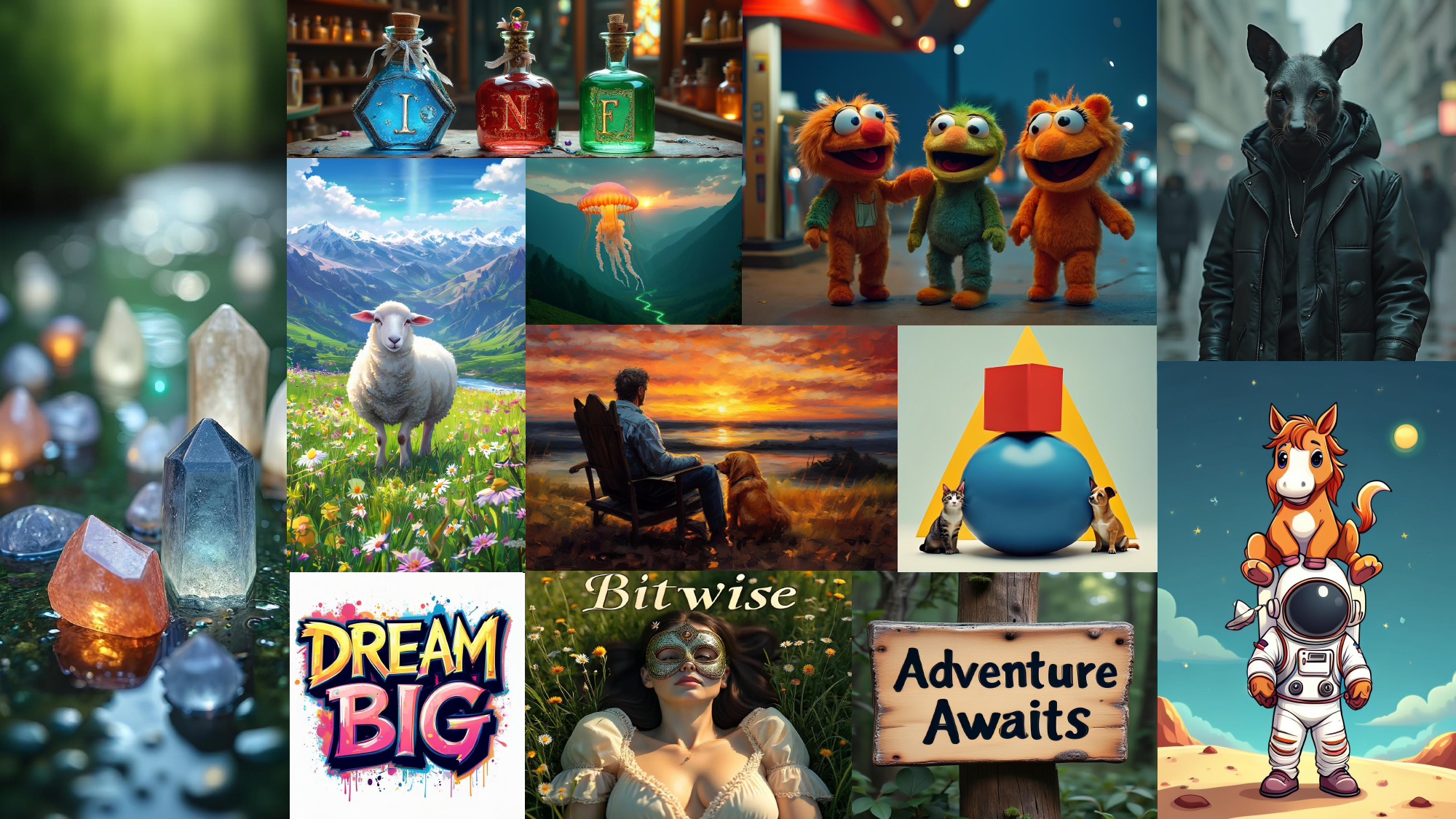}
\end{center}
\vspace{-7pt}
\caption{\small
\textbf{High-resolution image synthesis results from \methodNAME}, showcasing its capabilities in precise prompt following, spatial reasoning, text rendering, and aesthetics across different styles and aspect ratios.
}
\label{fig:abs_fig}
\end{figure}

\begin{abstract}
\vspace{-0.2cm}
We present \methodNAME, a Bitwise Visual AutoRegressive Modeling capable of generating high-resolution, photorealistic images following language instruction.  \methodNAME redefines visual autoregressive model under a bitwise token prediction framework with an infinite-vocabulary tokenizer \& classifier and bitwise self-correction mechanism, remarkably improving the generation capacity and details. By theoretically scaling the tokenizer vocabulary size to infinity and concurrently scaling the transformer size, our method significantly unleashes powerful scaling capabilities compared to vanilla VAR. 
\methodNAME sets a new record for autoregressive text-to-image models, outperforming top-tier diffusion models like SD3-Medium and SDXL. Notably, \methodNAME surpasses SD3-Medium by improving the GenEval benchmark score from \emph{0.62} to \emph{0.73} and the ImageReward benchmark score from \emph{0.87} to \emph{0.96}, achieving a win rate of \emph{66\%}. Without extra optimization, \methodNAME generates a high-quality \emph{1024}$\times$\emph{1024} image in 0.8 seconds, making it \emph{2.6}$\times$ faster than SD3-Medium and establishing it as the fastest text-to-image model. Models and codes will be released to promote further exploration of \methodNAME for visual generation and unified tokenizer modeling.

\end{abstract}

\begin{figure}[t]
  \centering
   \includegraphics[width=0.832\linewidth]{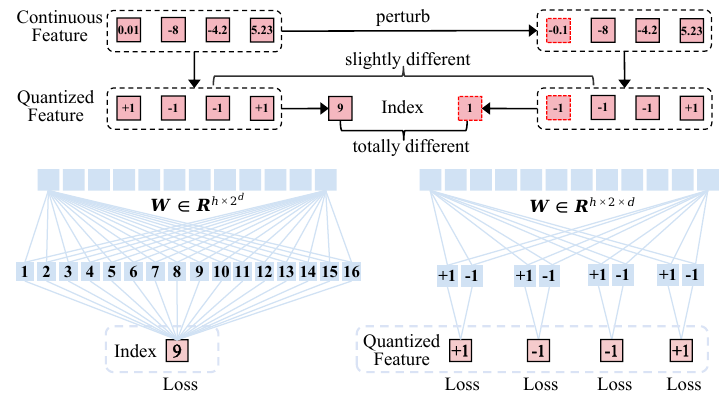}
   \caption{Visual tokenizer quantizes continuous features and then gets index labels. Conventional classifier (left) predicts $2^d$ indices. Infinite-Vocabulary Classifier (right) predicts $d$ bits instead. Slight perturbations to near-zero values in continuous features cause a complete change of index labels. Bit labels (\emph{i.e.} quantized features) change subtly and still provide steady supervision. Besides, parameters of conventional classifiers grow exponentially as $d$ increases, while IVC grows linearly. If $d = 32$ and $h=2048$, the conventional classifier requires \textbf{8.8 trillion} parameters,  exceeding current compute limits. By contrast, IVC only requires \textbf{0.13M} parameters.}
   \label{fig:ivc}
\end{figure}

\section{Introduction}
\label{sec:intro}

The visual generation\cite{ddpm, ddim, adm, ldm, dit} has recently witnessed rapid advancements, enabling high-quality, high-resolution images and video synthesis\cite{sora, stable-diffusion3}. Text-to-image generation\cite{imagen, dalle1, dalle2, DALLE3, sdxl, stable-diffusion3} is one of the most challenging tasks due to its need for complex language adherence and intricate scene creation. Currently, visual generation is primarily divided into two main approaches: Diffusion models and AutoRegressive models.  

Diffusion models\cite{ddpm, ddim, adm, sdxl, dit, stable-diffusion3}, trained to invert the forward paths of data towards random noise, effectively generate images through a continuous denoising process. AutoRegressive models\cite{igpt,vqgan,parti,VAR}, on the other hand, harness the scalability and generalizability of language models\cite{palm,palm2,chinchilla,llama1,llama2,bloom,ernie3,qwen,team2023gemini} by employing a visual tokenizer\cite{vqvae, vqvae2, vit-vqgan} to convert images into discrete tokens and optimize these tokens causally, allowing image generation through next-token prediction or next-scale prediction.
AutoRegressive models encounter significant challenges in high-resolution text-to-image synthesis\cite{parti, wang2024emu3}. They exhibit inferior reconstruction quality when utilizing discrete tokens as opposed to continuous tokens. Additionally, their generated visual contents are less detailed than those by diffusion models. The inefficiency and latency in visual generation, stemming from the raster-scan method of next-token prediction, further exacerbates these issues.

Recently, Visual AutoRegressive modeling (VAR)\cite{VAR} redefined autoregressive learning on images as coarse-to-fine “next-scale prediction”. It demonstrates superior generalization and scaling capabilities compared to diffusion transformers while requiring fewer steps. VAR leverages the powerful scaling properties of LLMs~\cite{scalinglaw,scalingar} and can simultaneously refine previous scale steps, benefiting from the strengths of diffusion models as well.
However, the index-wise discrete tokenizer\cite{vqvae, vqgan, titok, VAR, imagefolder, omnitokenizer, maskbit, titok} employed in AutoRegressive or Visual AutoRegressive models faces significant quantization errors with a limited vocabulary size resulting in difficulties in reconstructing fine-grained details especially in high-resolution images. In the generation stage, index-wise tokens suffer from fuzzy supervision leading to visual detail loss and local distortions. Moreover, train-test discrepancies from teacher-forcing training, inherent to LLMs, amplify cumulative errors in visual details. These challenges make index-wise tokens a significant bottleneck for AutoRegressive models.

We propose a novel approach called bitwise modeling, which substitutes index-wise tokens with bitwise tokens throughout the process. Our bitwise modeling framework consists of three primary modules:  bitwise visual tokenizer, bitwise infinite-vocabulary classifier, and bitwise self-correction. 
Inspired by the success and widespread adoption of binary vector quantization\cite{LFQ, BSQ}, we scaled up the tokenizer vocabulary to $2^{64}$,  significantly surpassing all previous AutoRegressive model vocabularies\cite{magvit2, llamagen}. This expansion allows for reconstruction quality that far exceeds previous discrete tokenizers, achieving results comparable to continuous VAEs\cite{ldm}, with scores improving from 0.87 to 0.33 on ImageNet-256 benchmark\cite{imagenet}. In Fig.\ref{fig:ivc}, we transform the conventional token prediction from a large integer into binary bits in a bitwise infinite-vocabulary classifier to address optimization and computation challenges, enabling the learning of massive vocabularies in Visual AutoRegressive models. Additionally, we incorporated bitwise self-correction during training by randomly flipping some bits to simulate prediction mistakes and re-quantizing the residual features, thus endowing the system with self-correcting capabilities. Our method, Infinity: Bitwise Visual AutoRegressive Modeling, maintains the scaling and speed advantages of Visual AutoRegressive modeling while achieving detailed reconstruction and generation quality comparable to that of continuous Diffusion models.

\methodNAME sets a new record for AutoRegressive models, and also surpasses leading diffusion models including SDXL\cite{sdxl}, PixArt-Sigma\cite{chen2024pixart_sigma},DALL-E3\cite{DALLE3} and Stable-Diffusion 3\cite{stable-diffusion3} on several challenging text-to-image benchmarks. Notably, \methodNAME surpasses SD3 by improving the GenEval benchmark score from \emph{0.62} to \emph{0.73}, ImageReward benchmark score from
\emph{0.87} to \emph{0.96}, HPSv2.1 benchmark score from \emph{30.9} to \emph{32.3}, achieving a win rate of \emph{66\%} for human evaluation and a \emph{2.6$\times$} reduction in inference latency with the same model size. Specifically, \methodNAME shows powerful scaling laws for image generation capabilities by scaling up the image tokenizer vocabulary size and the corresponding transformer size. As the image tokenizer and transformer sizes increase, the content and details of high-quality image generation show significant improvement.

In summary, the contributions of our work are as follows:
\begin{enumerate} 
\item  We propose \methodNAME, an autoregressive model with Bitwise Modeling, which significantly improves the scaling and visual detail representation capabilities of discrete generative models. We believe this framework opens up new possibilities of `infinity' for the discrete generation community.

\item \methodNAME demonstrates the potential of scaling tokenizers and transformers by achieving near-continuous tokenizer performance with its image tokenizer and surpassing diffusion models in high-quality text-to-image generation.

\item  \methodNAME enables a discrete autoregressive text-to-image model to achieve exceptionally strong prompt adherence and superior image generation quality, while also delivering the fastest inference speed.
\end{enumerate}

\section{Related Work}
\label{sec:related_works}

\subsection{AutoRegressive Models}
AutoRegressive models, leveraging the powerful scaling capabilities of LLMs\cite{gpt2,gpt3,palm,llama1,llama2}, use discrete image tokenizers\cite{vqvae, vqvae2, vqgan} in conjunction with transformers to generate images based on next-token prediction. VQ-based methods~\cite{vqvae,vqvae2, vqgan,rq,llamagen} employ vector quantization to convert image patches into index-wise tokens and use a decoder-only transformer to predict the next token index. However, these methods are limited by the lack of scaled-up transformers and the quantization error inherent in VQ-VAE\cite{vqvae}, preventing them from achieving performance on par with diffusion models.  Parti~\cite{parti}, Emu3~\cite{wang2024emu3}, chameleon\cite{chameleon-meta}, loong\cite{loong} and VideoPoet\cite{videopoet} scaled up autoregressive models in text-to-image or video synthesis. Inspired by the global structure of visual information, Visual AutoRegressive modeling(VAR) redefines the autoregressive modeling on images as a next-scale prediction framework, significantly improving generation quality and sampling speed. HART\cite{tang2024hart} adopted hybrid tokenizers based on VAR. Fluid\cite{Fluid} proposed random-order models and employed a continuous tokenizer rather than a discrete tokenizer.

\subsection{Diffusion Models.}
Diffusion models have seen rapid advancements in various directions. Denoising learning mechanisms~\cite{ddpm,nichol2021improved} and sampling efficiency~\cite{scorebased,ddim,dpm-solver,dpmpp,bao2022analytic} have been continuously optimized to generate high-quality images. Latent diffusion models~\cite{ldm} is the first to propose modeling in the latent space rather than the pixel space for diffusion\cite{imagen}. Recently, latent diffusion models\cite{dai2023emu_meta, stable-diffusion3} have also adopted scaling up VAE to improve the representation in the latent space. DiT~\cite{dit} and U-Vit\cite{bao2023all} employ a more scalable transformer to model diffusion, achieving superior results. Consequently, mainstream text-to-image diffusion models\cite{stable-diffusion3,DALLE3,chen2023pixart} have adopted the DiT architecture. DiT also inspire the text-to-video diffusion models\cite{bao2024vidu, sora}

\subsection{Scaling models}
Scaling laws in autoregressive language models reveal a power-law relationship between model size, dataset size, and compute with test set cross-entropy loss \cite{scalinglaw, scalingar, gpt4}. These laws help predict larger model performance, leading to efficient resource allocation and ongoing improvements without saturation \cite{gpt3, llama1, llama2, opt, bloom, chinchilla}. This has inspired research into scaling in visual generation
\cite{emu_baai, cm3leon_chameleon, VAR, stable-diffusion3, sora}

\begin{figure}[t]
  \centering
   \includegraphics[width=1.0\linewidth]{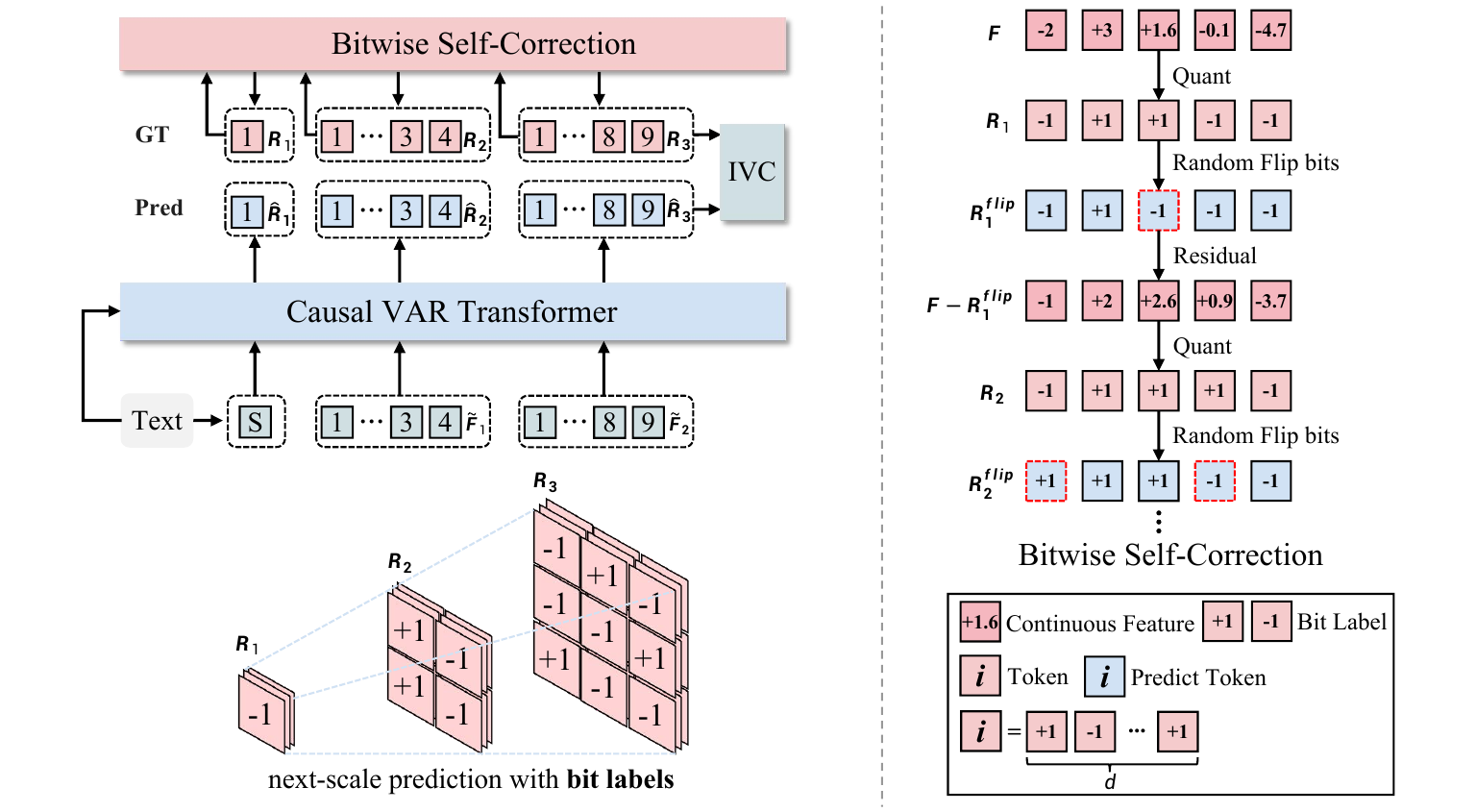}
   \caption{\textbf{Framework of \methodNAME.} \methodNAME introduces bitwise modeling, which incorporates a bitwise multi-scale visual tokenizer, Infinite-Vocabulary Classifier (IVC), and Bitwise Self-Correction. When predicting $\bm{R}_k$, the sequence $(\bm{R}_1, {\bm{R}}_2, ..., \bm{R}_{k-1})$ serves as the prefixed context and the text condition guides the prediction through a cross attention mechanism. Different from VAR, \methodNAME performs next-scale prediction with bit labels.}
   \label{fig:framework}
\end{figure}

\section{\methodNAME Architecture}

\subsection{Visual AutoRegressive Modeling}

\methodNAME incorporates a visual tokenizer and a transformer for image synthesis. During the training stage, a sample consists of a text prompt $t$ and a ground truth image $\bm{I}$. The proposed visual tokenizer first encodes the image $\bm{I}$ into a feature map $\bm{F}\in \mathbb{R}^{h \times w \times d}$ with stride $s$ and then quantize the feature map $\bm{F}$ into $K$ multi-scale residual maps $(\bm{R}_1, \bm{R}_2, ..., \bm{R}_K)$.
The resolution of $\bm{R}_k$ is $h_k \times w_k$ and it grows larger gradually from $k=1 \to K$. 
Based on this sequence of residuals, we can gradually approximate the continuous feature $\bm{F}$ as in Eq.\ref{eq:cum_sum} 
\begin{equation}
    \bm{F}_k = \sum_{i=1}^{k} \operatorname{up}(\bm{R}_i, (h, w))
  \label{eq:cum_sum}
\end{equation}
Here $\operatorname{up}(\cdot)$ means bilinear upsampling and $\bm{F}_k$ is the cumulative sum of the upsampled $\bm{R}_{\leq k}$.

Subsequently, transformer learns to predict residuals $\bm{R}$ of the next scale conditioned on previous predictions and the text input in an autoregressive manner. Formally, the autoregressive likelihood can be formulated as: 
\begin{equation}
p(\bm{R}_1,...,\bm{R}_K)=\prod_{k=1}^{K} p(\bm{R}_k \mid \bm{R}_1,...,\bm{R}_{k-1},\bm{\Psi}(t)),
  \label{eq:next_scale_predict}
\end{equation}
where $\bm{\Psi}(t)$ is the text embeddings from Flan-T5~\cite{Flan-T5} model. $(\bm{R}_1, ..., \bm{R}_{k-1}, \bm{\Psi}(t))$ serves as the prefixed context When predicting $\bm{R}_k$. Besides, the text embeddings $\bm{\Psi}(t)$ further guide the prediction through a cross attention mechanism. In particular, as shown in Fig.~\ref{fig:framework}, the text embeddings $\bm{\Psi}(t) \in \mathbb{R}^{L \times C}$ is projected into a $\bm{\langle \text{SOS} \rangle} \in \mathbb{R}^{1\times 1 \times h}$ as the input of the first scale, where $h$ is the hidden dimension of transformer. The transformer is required to predict $\bm{R}_1$ based on $\bm{\langle \text{SOS} \rangle}$ in the first scale. In the latter $k$-th scale, to match the spatial size of the input and the output label $\bm{R}_k$, we take the downsampled feature $\widetilde{\bm{F}}_{k-1}$ from the last scale $k-1$ as the input to predict $\bm{R}_k$ in parallel. 
\begin{equation}
    \widetilde{\bm{F}}_{k-1} = \operatorname{down}(\bm{F}_{k-1}, (h_{k}, w_{k})),
  \label{eq:tide_fk}
\end{equation}
where $\operatorname{down}(\cdot)$ is bilinear downsampling and the spatial size of both $\widetilde{\bm{F}}_{k-1}$ and $\bm{R}_k$ is $(h_k, w_k)$. Refer to Alg.\ref{alg:vae} for detailed procedure to obtain binary quantization results and transformer's inputs. In previous index-wise~\cite{VAR} representations, the shape of prediction is $(h_k, w_k, V_d)$. $V_{d}$ is the vocabulary size of the visual tokenizer. For binary quantization~\cite{LFQ,BSQ} with code embedding dimension $d$, $V_d=2^d$. When $d$ is large, the required computational resource grows unaffordable.

The transformer consists of a stack of repeated blocks, where each block includes RoPE2d~\cite{ROPE2d}, Self-Attention, Cross Attention, and FFN layers. The text embeddings $\bm{\Psi}(t)$ provide effective guidance for image synthesis in each cross-attention layer. During the training stage, we exploit a block-wise causal attention mask to ensure that the transformer can only attend to its prefixed context, \textit{i.e.}, $(\bm{\langle \text{SOS} \rangle}, \widetilde{\bm{F}}_1, ..., \widetilde{\bm{F}}_{k-1})$, when predicting $\bm{F}_k$. During the inference stage, we perform KV-Caching to speed up inference and there's no need for masking.

\subsection{Visual Tokenizer}
\label{subsec:vae}
Increasing the vocabulary size has significant potential for improving reconstruction and generation quality. However, directly enlarging the vocabulary in existing tokenizers\cite{magvit2,VAR} leads to a substantial increase in memory consumption and computational burden. 
To address these challenges and fully exploit the potential of discrete tokenizers, this paper proposes a new \textbf{bitwise multi-scale residual quantizer}, which significantly reduces memory usage, enabling the training of extremely large vocabulary, e.g. $2^{64}$. 

\textbf{Bitwise Multi-scale Residual Quantizer}. We replace the original vector quantizer of VAR~\cite{VAR} with a dimension-independent bitwise 
quantizer. In this paper, we consider two candidates, LFQ~\cite{magvit2} and BSQ\cite{BSQ}. 
Given $K$ scales in the multi-scale quantizer, on the $k$-th scale, the input continuous residual vector $z_k\in \mathbb{R}^{d}$ are quantized into binary output $q_k$ as shown below.
\begin{equation}
q_k = \mathcal{Q}(z_k) =
\begin{cases} 
\mathrm{sign}(z_k) & \text{if } \mathrm{LFQ} \\
\frac{1}{\sqrt d} \mathrm{sign}(\frac{z_k}{|z_k|}) & \text{if } \mathrm{BSQ}
\end{cases}
\label{eq:vae-quant}
\end{equation}

To encourage codebook utilization, an entropy penalty $\mathcal{L}_{entropy}=\mathbb{E}[H(q(z))]-H[\mathbb{E}(q(z))]$ \cite{entropy-penalty} is adopted, where $H(\cdot)$ represents the entropy operation. To obtain the distribution of $q(z)$, we need to compute the similarities between the input $z$ and the whole codebook when using LFQ. However, this leads to unaffordable space and time complexity of $O(2^d)$. When the codebook dimension $d$ becomes large, e.g. 20, an out-of-memory (OOM) issue is faced as shown in Tab.~\ref{tab:memory-comparison}. By contrast, since both input and output in BSQ are unit vectors, BSQ\cite{BSQ} proposes an approximation formula for the entropy penalty, reducing the computational complexity to $O(d)$. As shown in Tab~\ref{tab:memory-comparison}, there is no obvious increase in memory consumption for BSQ even when codebook size is $2^{64}$. Unless otherwise stated, we adopt BSQ by default.

\subsection{Infinite-Vocabulary Classifier}
The visual tokenizer obtains discrete labels by quantizing residual features. Consequently, the transformer predicts next-scale residual features' labels $\bm{y}_k \in [0,V_d)^{h_k \times w_k}$ and optimizes the target through the cross-entropy loss. Previous works~\cite{VAR,LFQ} directly predict these index labels using a classifier of $V_d$ classes. However, it suffers from two drawbacks, huge computational costs and fuzzy supervision. 

As illustrated in Section \ref{subsec:vae}, we exploit a bitwise VQ-VAE as the visual tokenizer, where the vocabulary size $V_d$ is extremely large. For example, if $V_d = 2^{32}$ and $h=2048$, a conventional classifier would require a weight matrix $W\in\mathbb{R}^{h\times V_d}$ of 
8.8 trillion parameters, which exceeds the limits of current computational resources.

Moreover, VQ-VAE exploits the sign function during quantization as in Eq.\ref{eq:vae-quant}. After that, the positive elements are multiplied with the corresponding base and summed to get the index label $\bm{y}_{k}(m,n)$ as in Eq.\ref{eq:label}, where $m \in [0,h_k)$ and $n \in [0,w_k)$.

\begin{equation}
\bm{y}_{k}(m,n)=\sum_{p=0}^{d-1} \mathbbm{1}_{\bm{R}_{k}(m,n,p)>0} \cdot 2^p
\label{eq:label}
\end{equation}
Owing to the merits of the quantization method, slight perturbations to those near-zero features cause a significant change in the label. As a result, the conventional index-wise classifier~\cite{VAR,maskgit,magvit2} is difficult to optimize. 

To address the problems in computation and optimization, we propose Infinite-Vocabulary Classifier (IVC). As shown in Fig.\ref{fig:ivc}, instead of using a conventional classifier with ${V_d}$ classes, we use $d$ binary classifiers in parallel to predict if the next-scale residual $\bm{R}_{k}(m,n,p)$ is positive or negative, where $d=log_2(V_d)$. The proposed Infinite-Vocabulary Classifier is much more efficient in memory and parameters compared to the conventional classifier. When $V_d=2^{16}$ and $h=2048$, it saves 99.95\% parameters and GPU memory. Besides, when there exist near-zero values that confuse the model in some dimensions, the supervision in other dimensions is still clear. Therefore, compared with conventional index-wise classifiers, the proposed Infinite-Vocabulary Classifier is easier to optimize.

\subsection{Bitwise Self-Correction}
\textbf{Weakness of teacher-forcing training.} VAR~\cite{VAR} inherits the teacher-forcing training from LLMs. However, next-scale prediction in vision is quite different from next-token prediction in language. Specifically, we cannot decode the complete image until residuals $\bm{R}_k$ from all scales are obtained. We find that the teacher-forcing training brings about severe train-test discrepancy for visual generation. In particular, the teacher-forcing training makes the transformer only refine features in each scale without the ability to recognize and correct mistakes. Mistakes made in former scales will be propagated and amplified in latter scales, finally messing up generated images (left images in Fig.\ref{fig:self_correction}). 

In this work, we propose Bitwise Self-Correction (BSC) to address this issue. In particular, we obtain ${\bm{R}^{flip}_k}$ via randomly flipping the bits in $\bm{R}_k$ with a probability uniformly sampled from $[0,p]$, imitating different strengths of errors made in the prediction of the $k$-th scale. 

Here comes the key component of bitwise self-correction. $\bm{R}^{flip}_{k}$ contains errors while $\bm{R}_{k}$ doesn't. After replacing $\bm{R}_{k}$ with $\bm{R}^{flip}_{k}$ as predictions on the $k$-th scale, 
we recompute the transformer input $\widetilde{\bm{F}}_k$. Besides, re-quantization is performed to get a new target $\bm{R}_{k+1}$. The whole process of bitwise self-correction is illustrated in Alg.\ref{alg:self_correction}. We also provide a simplified illustration in Fig.\ref{fig:framework} (right) for better understanding. Notably, BSC is accomplished by revising the inputs and labels of the transformer. It neither adds extra computational cost nor disrupts the original parallel training characteristics.

Each scale undergoes the same process of random-flipping and re-quantization. The transformer takes partially randomly flipped features as inputs, taking the prediction errors into consideration. The re-quantized bit labels enable the transformer to autocorrect errors made in former predictions. In such way, we address the train-test discrepancy caused by teacher-forcing training and empower \methodNAME to have the self-correction ability.

\renewcommand{\algorithmicrequire}{\textbf{Input:}}
\renewcommand{\algorithmicensure}{\textbf{Output:}}

\begin{center}
\begin{minipage}[t]{0.46\linewidth}
\begin{algorithm}[H]
\caption{Visual Tokenizer Encoding} \label{alg:vae}
\begin{algorithmic}[0]
\Require raw feature $\bm{F}$, scale schedule $\{(h^r_1,w^r_1),...,(h^r_K,w^r_K)\}$
\State $\bm{R}_{queue}=[]$ \Comment{multi-scale bit labels}
\State $\widetilde{\bm{F}}_{queue}=[]$ \Comment{inputs for transformer}
\For {$k=1,2,\cdots,K \vphantom{\bm{F}^{flip}_{k-1}}$}
\State $\bm{R}_k=\mathcal{Q}(\operatorname{down}(\bm{F} - \bm{F}_{k-1}, (h_{k}, w_{k}))$
\State $\operatorname{Queue\_Push}$($\bm{R}_{queue}, \bm{R}_{k}$)
\State $\bm{F}_{k} = \sum_{i=1}^{k} \operatorname{up}(\bm{R}_i, (h, w))$
\State $\widetilde{\bm{F}}_k = \operatorname{down}(\bm{F}_{k}, (h_{k+1}, w_{k+1}))$
\State $\operatorname{Queue\_Push}$($\widetilde{\bm{F}}_{queue}, \widetilde{\bm{F}}_k$)
\EndFor
\Ensure $\bm{R}_{queue}, \widetilde{\bm{F}}_{queue}$ 
\end{algorithmic}
\end{algorithm}
\end{minipage}
\hfill
\begin{minipage}[t]{0.47\linewidth}
\raggedright
\begin{algorithm}[H]
\caption{Encoding with BSC} \label{alg:self_correction}
\begin{algorithmic}[0]
\Require raw feature $\bm{F}$, random flip ratio $p$, scale schedule $\{(h^r_1,w^r_1),...,(h^r_K,w^r_K)\}$, 
\State $\bm{R}_{queue}=[]$, $\widetilde{\bm{F}}_{queue}=[]$
\For {$k=1,2,\cdots,K$}
\State $\bm{R}_{k}$=$\mathcal{Q}(\operatorname{down}(\bm{F} - \bm{F}^{flip}_{k-1}, (h_{k}, w_{k})))$
\State $\operatorname{Queue\_Push}$($\bm{R}_{queue}, \bm{R}_{k}$)
\State $\bm{R}^{flip}_{k} = \operatorname{Random\_Flip}(\bm{R}_{k}, p)$
\State $\bm{F}^{flip}_{k} = \sum_{i=1}^{k} \operatorname{up}(\bm{R}^{flip}_i, (h, w))$
\State $\widetilde{\bm{F}}_k = \operatorname{down}(\bm{F}^{flip}_{k}, (h_{k+1}, w_{k+1}))$
\State $\operatorname{Queue\_Push}$($\widetilde{\bm{F}}_{queue}, \widetilde{\bm{F}}_k$)
\EndFor
\Ensure $\bm{R}_{queue}, \widetilde{\bm{F}}_{queue}$
\end{algorithmic}
\end{algorithm}
\end{minipage}
\end{center}

\subsection{Dynamic Aspect Ratios and Position Encoding}

\methodNAME can generate photo-realistic images with various aspect ratios, which is significantly different from VAR~\cite{VAR} that can only generate square images. The main obstacles of generating various aspect ratio images lie in two folds. The first is to define the height $h_k$ and width $w_k$ of $\bm{R}_k$ based on varying aspect ratios. In the supplementary material, we pre-define a list of scales, also called scale schedule, as $\{(h^r_1,w^r_1),...,(h^r_K,w^r_K)\}$ for each aspect ratio. We 
 ensure that the aspect ratio of each tuple $(h^r_k,w^r_k)$ is approximately equal to $r$, especially in the latter prediction scales. Additionally, for different aspect ratios at the same scale $k$, we keep the area of $h^r_k \times w^r_k$ to be roughly equal, ensuring that the training sequence lengths are roughly the same. 
 
Secondly, we need to carefully design a resolution-aware positional encoding method to handle features of various scales and aspect ratios. This issue poses a significant challenge, as the existing solutions~\cite{transformer,VAR,ROPE,ROPE2d,STAR} exhibit substantial limitations under such conditions. In this paper, we apply RoPE2d~\cite{ROPE2d} on features of each scale to preserve the intrinsic 2D structure of images. Additionally, we exploit learnable scale embeddings to avoid confusion between features of different scales. Compared to learnable APE element-wisely applied on features, learnable embeddings applied on scales bring fewer parameters, can adapt to varying sequence lengths, and are easier to optimize.

\section{Experiment}
\label{sec:experiment}

\subsection{Dataset}

\textbf{Data Curation.} We curated a large-scale dataset from open-source academic data and high-quality internally collected data.
The pre-training dataset is constructed by collecting and cleaning open-source academic datasets such as LAION \cite{schuhmann2021laion}, COYO \cite{kakaobrain2022coyo-700m}, OpenImages \cite{kuznetsova2020open}. We exploit an OCR model and a watermark detection model to filter undesired images with too many texts or watermarks. Additionally, we employ Aesthetic-V2 to filter out images with low aesthetic scores.

\begin{table}[t]
\centering
\captionsetup{skip=5pt} 
\caption{Evaluation on the GenEval~\cite{ghosh2024geneval} and DPG~\cite{DPG-bench} benchmark. $\dagger$ result is with prompt rewriting.}\label{tab:genEvalSotaTable}
\resizebox{\linewidth}{!}{ 
\begin{tabular}{lcccccccc}
        \toprule
    	\multirow{2}{*}{Methods} & \multirow{2}{*}{\# Params} & \multicolumn{4}{c}{GenEval$\uparrow$} & \multicolumn{3}{c}{DPG$\uparrow$} \\\cmidrule(l){3-6}\cmidrule(l){7-9}
        & & Two Obj. & Position & Color Attri. & \textbf{Overall} & Global & Relation & \textbf{Overall} \\
    	\midrule
            \multicolumn{9}{l}{Diffusion Models} \\
            \midrule
            LDM~\cite{rombach2022high} & 1.4B & 0.29 & 0.02 & 0.05 & 0.37 & - & - & - \\
            SDv1.5~\cite{rombach2022high} & 0.9B  & 0.38 & 0.04 & 0.06 & 0.43 &  74.63  & 73.49  &   63.18 \\
            PixArt-alpha~\cite{Pixart-alpha} & 0.6B & 0.50 & 0.08 & 0.07 & 0.48 &  74.97  & 82.57  &   71.11 \\
            SDv2.1~\cite{rombach2022high} & 0.9B & 0.51 & 0.07 & 0.17 & 0.50 & 77.67   & 80.72 &   68.09 \\
            DALL-E 2~\cite{dalle2}& 6.5B & 0.66 & 0.10 & 0.19 & 0.52 & - & - & - \\
            DALL-E 3~\cite{DALLE3}& - & - & - & - & 0.67$^{\dagger}$ &  90.97  & 90.58  &   83.50 \\
            SDXL~\cite{sdxl} & 2.6B & 0.74 & 0.15 & 0.23 &  0.55 &    83.27   &   86.76   & 74.65 \\
            PixArt-Sigma~\cite{chen2024pixart_sigma} & 0.6B & 0.62 & 0.14 & 0.27 & 0.55 &   86.89   & 86.59 &   80.54 \\
            SD3 (d=24)~\cite{stable-diffusion3} & 2B & 0.74 & 0.34 & 0.36 & 0.62 & - & - & 84.08 \\
            SD3 (d=38)~\cite{stable-diffusion3} & 8B  &  0.89 & 0.34 & 0.47 & 0.71 & - & - &  - \\
            \midrule
            \multicolumn{9}{l}{AutoRegressive Models} \\
            \midrule
            LlamaGen~\cite{llamagen} & 0.8B & 0.34 & 0.07 & 0.04 & 0.32 &  &   &   65.16 \\
            Chameleon~\cite{chameleon-meta} & 7B & - & - & - & 0.39 & - &- & - \\
            HART~\cite{tang2024hart} & 732M & - & - & - & 0.56 & - &- & 80.89 \\
            Show-o~\cite{show-o} & 1.3B & 0.80 &  0.31 &  0.50 & 0.68 & - &- & 67.48 \\
            Emu3~\cite{wang2024emu3} & 8.5B & 0.81$^{\dagger}$ & 0.49$^{\dagger}$ & 0.45$^{\dagger}$ & 0.66$^{\dagger}$ & - & - & 81.60 \\
            \cellcolor{mycolor_green}{\textbf{\methodNAME}} & \cellcolor{mycolor_green}{2B} & \cellcolor{mycolor_green}{0.85$^{\dagger}$} & \cellcolor{mycolor_green}{\textbf{0.49}$^{\dagger}$} & \cellcolor{mycolor_green}{\textbf{0.57}$^{\dagger}$} & \cellcolor{mycolor_green}{\textbf{0.73}$^{\dagger}$} & \cellcolor{mycolor_green}{\textbf{93.11}} & \cellcolor{mycolor_green}{\textbf{90.76}} & \cellcolor{mycolor_green}{83.46}\\
        \bottomrule
\end{tabular}
}
\end{table}

\subsection{Implementation}
\methodNAME redefines text-to-image as a coarse-to-fine, next-scale prediction task. In line with its architecture, we propose to train \methodNAME in a progressive strategy. Specifically, we first train \methodNAME of 2B parameters on the pre-training dataset with 256 resolution for 150k iterations using a batch size of 4096 and a learning rate of 6e-5. Then we switch to 512 resolution and train 110k iterations using the same hyper-parameters. Next, we fine-tune \methodNAME at 1024 resolution with a smaller, high-quality dataset. In this stage, we train \methodNAME for 60k iterations using a batch size of 2048 and a learning rate of 2e-5. All training stages use images with varying aspect ratios.

\begin{figure}[H]
  \centering
   \includegraphics[width=1.0\linewidth]{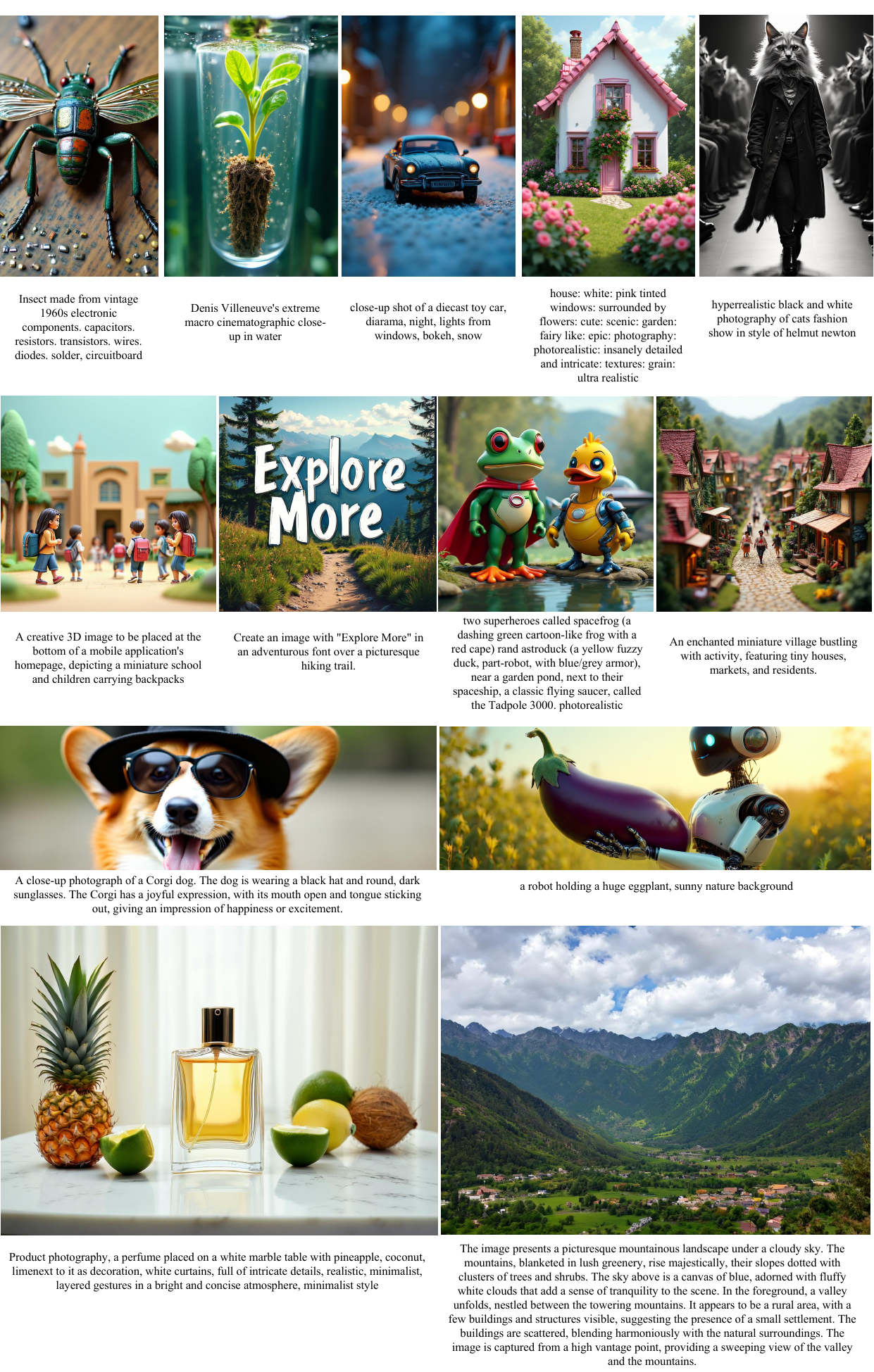}
   \caption{Qualitative results from \methodNAME.}
   \label{fig:show_images}
\end{figure}

As for evaluation, we report results on popular text-to-image benchmarks like GenEval~\cite{ghosh2024geneval} and DPG~\cite{DPG-bench}. We also measure our method on two human preference evaluation benchmarks, \emph{i.e.}, ImageReward \cite{xu2024imagereward} and HPSv2.1 \cite{wu2023human}. These two benchmarks have trained models to predict human preference scores by learning from abundant human-ranked text-image pairs. We also build a validation set consisting of 40K text-image pairs to measure FID.

\subsection{Text-to-Image Generation}
\textbf{Overall Results.} Fig.\ref{fig:abs_fig} and Fig.\ref{fig:show_images} present generated images from our \methodNAME-2B model, showcasing \methodNAME 's strong capabilities in generating high-fidelity images from various categories following user prompts. Qualitative comparison results among \methodNAME and other top-tier models can be found in the appendix. 

\noindent\textbf{Prompt-Following.}
Fig.\ref{fig:prompt_following_comparison} presents three examples demonstrating the superior prompt-following ability of \methodNAME. As highlighted in red, \methodNAME consistently adheres to user prompts, whether they are short or extremely long texts. We attribute these improvements to the bitwise token prediction and scaling autoregressive modeling.

\noindent\textbf{Text Rendering.}
As illustrated in Fig.\ref{fig:text_rendering}, \methodNAME can render text according to user prompts across diverse categories. Despite diverse backgrounds and subjects, \methodNAME accurately renders corresponding texts according to user requirements, such as fonts, styles, colors, and more. 

\textbf{Benchmark.} As in Tab \ref{tab:genEvalSotaTable}, on GenEval\cite{ghosh2024geneval}, our model with a re-writer achieves the best overall score of 0.73. Besides, \methodNAME also reaches the highest position reasoning score of 0.49.
On DPG~\cite{DPG-bench}. Our model reaches an overall score of 83.46, surpassing SDXL~\cite{sdxl}, Playground v2.5~\cite{Playground-2.5}, and DALLE 3~\cite{DALLE3}. What's more, \methodNAME achieves the best relation score of 90.76 among all open-source T2I models, demonstrating its stronger ability to generate spatially consistent images based on user prompts.

\newcommand{\dpgSotaTable}{
\begin{table}[t]
\tiny
\centering
\setlength{\tabcolsep}{2pt}
\resizebox{.99\linewidth}{!}{
\begin{tabular}{@{}lcccccccc@{}}
\toprule
Method & \# Params & Overall & Global & Entity & Attribute & Relation & Other \\
\midrule
Diffusion-based & & & & & &  \\
\midrule
SDXL~\cite{sdxl} & - & 74.65 & 83.27 & 82.43 & 80.91 & 86.76 & 80.41 \\
Playground v2.5~\cite{Playground-2.5} & 2B  & 75.47 & 83.06 & 82.59 & 81.20 & 84.08 & 83.50 \\
Lumina-Next~\cite{lumina-next} & - & 74.63 & 82.82 & 88.65 & 86.44 & 80.53 & 81.82 \\
Hunyuan-DiT~\cite{hunyuan-dit} & - & 78.87 & 84.59 & 80.59 & 88.01 & 74.36 & 86.41 \\
PixArt-Sigma~\cite{chen2024pixart_sigma} & 0.6B & 80.54 & 86.89 & 82.89 & 88.94 & 86.59 & 87.68 \\
DALLE 3~\cite{DALLE3} & - & 83.50 & 90.97 & 89.61 & 88.39 & 90.58 & 89.83 \\
SD3-Medium~\cite{stable-diffusion3} & 2B & 84.08 & 87.90 & 91.01 & 88.83 & 80.70 & 88.68 \\
\midrule
AutoRegressive-based & & & & & &  \\
\midrule
Emu3 & 2B & 81.60 & - &  - &  - &  - &  - \\
HART & 2B & 81.60 & - &  - &  - &  - &  - \\
\textbf{\methodNAME} & 2B & 83.46 & 93.11 & 89.45 & 88.40 & 90.76 & 80.51 \\
\bottomrule
\end{tabular}
}
\caption{\textbf{DPG-bench~\cite{DPG-bench} evaluation}. We compared our method with SOTA open-source models. From the table, our model achieved high text-image alignment performance.
}\label{tab:dpgSotaTable}
\end{table}
}

\begin{figure}[H]
  \centering
   \includegraphics[width=0.8\linewidth]{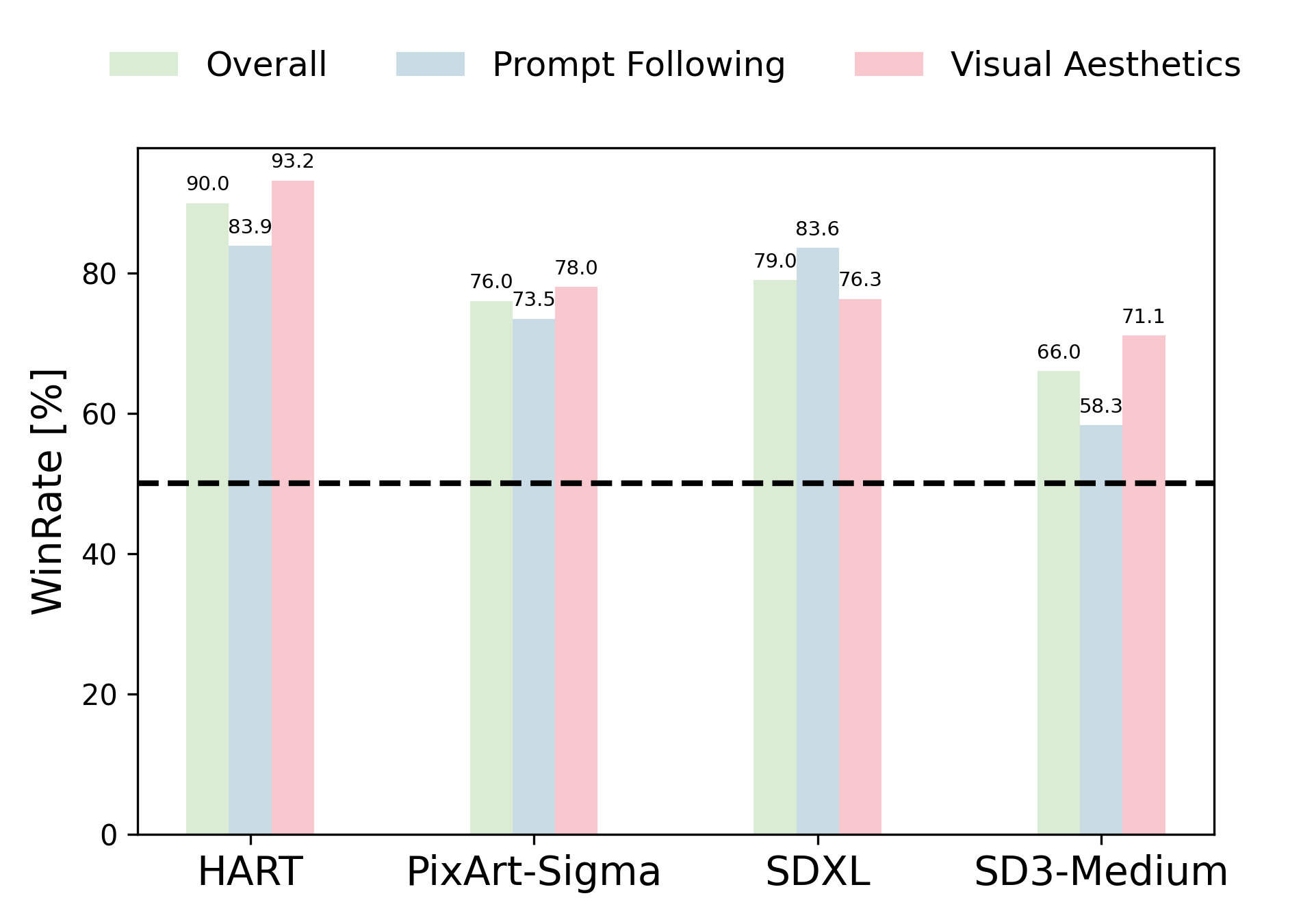}
   \caption{Human Preference Evaluation. We ask users to select the better one in a side-by-side comparison in terms of Overall Quality, Prompt Following, and Visual Aesthetics. \methodNAME is more preferred by humans compared to other open-source models.}
   \label{fig:human_eval}
\end{figure}

\noindent\textbf{Human Preference Evaluation.} 
We conduct human preference evaluation in both human studies and benchmarks. As in Fig.\ref{fig:human_eval}, the generation results of \methodNAME are more frequently selected by humans in terms of \emph{overall quality, prompt following, and visual aesthetics} in contrast to other open-sourced T2I models. Please refer to the appendix for more details. Tab.\ref{tab:hpsTable} lists the results of two human preference benchmarks, \textit{i.e.}, ImageReward \cite{xu2024imagereward} and HPSv2.1 \cite{wu2023human}. \methodNAME reaches the highest ImageReward and HPSv2.1, indicating our method could generate images that are more appealing to humans.

\begin{figure}[H]
  \centering
   \includegraphics[width=0.86\linewidth]{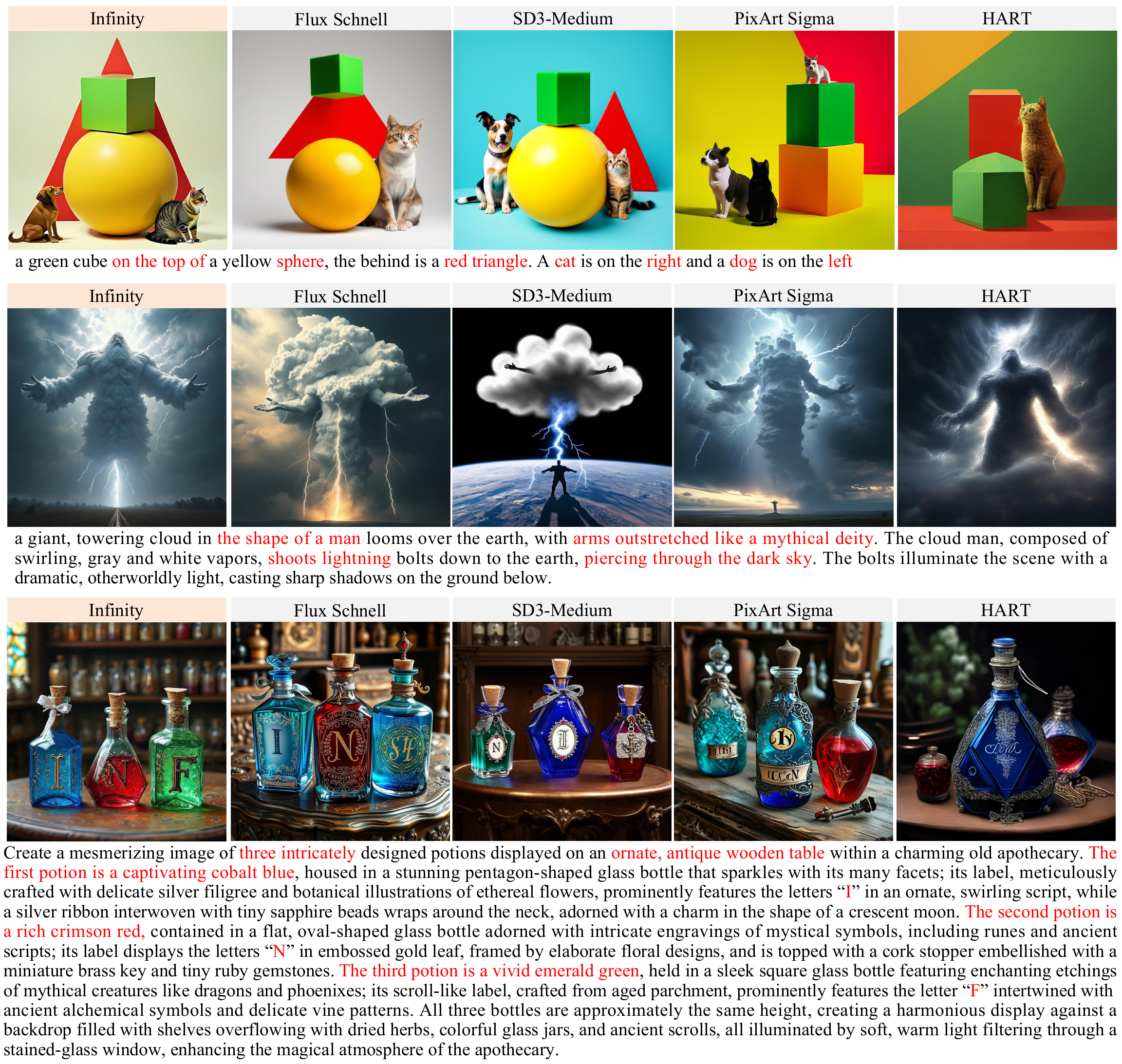}
   \vspace{-0.15cm}
   \caption{Prompt-following qualitative comparison. We highlight text in red that \methodNAME-2B consistently adheres to while the other four models fail to follow. Zoom in for better comparison.}
   \label{fig:prompt_following_comparison}
\end{figure}

\vspace{-0.45cm}
\begin{figure}[H]
  \centering
   \includegraphics[width=0.86\linewidth]{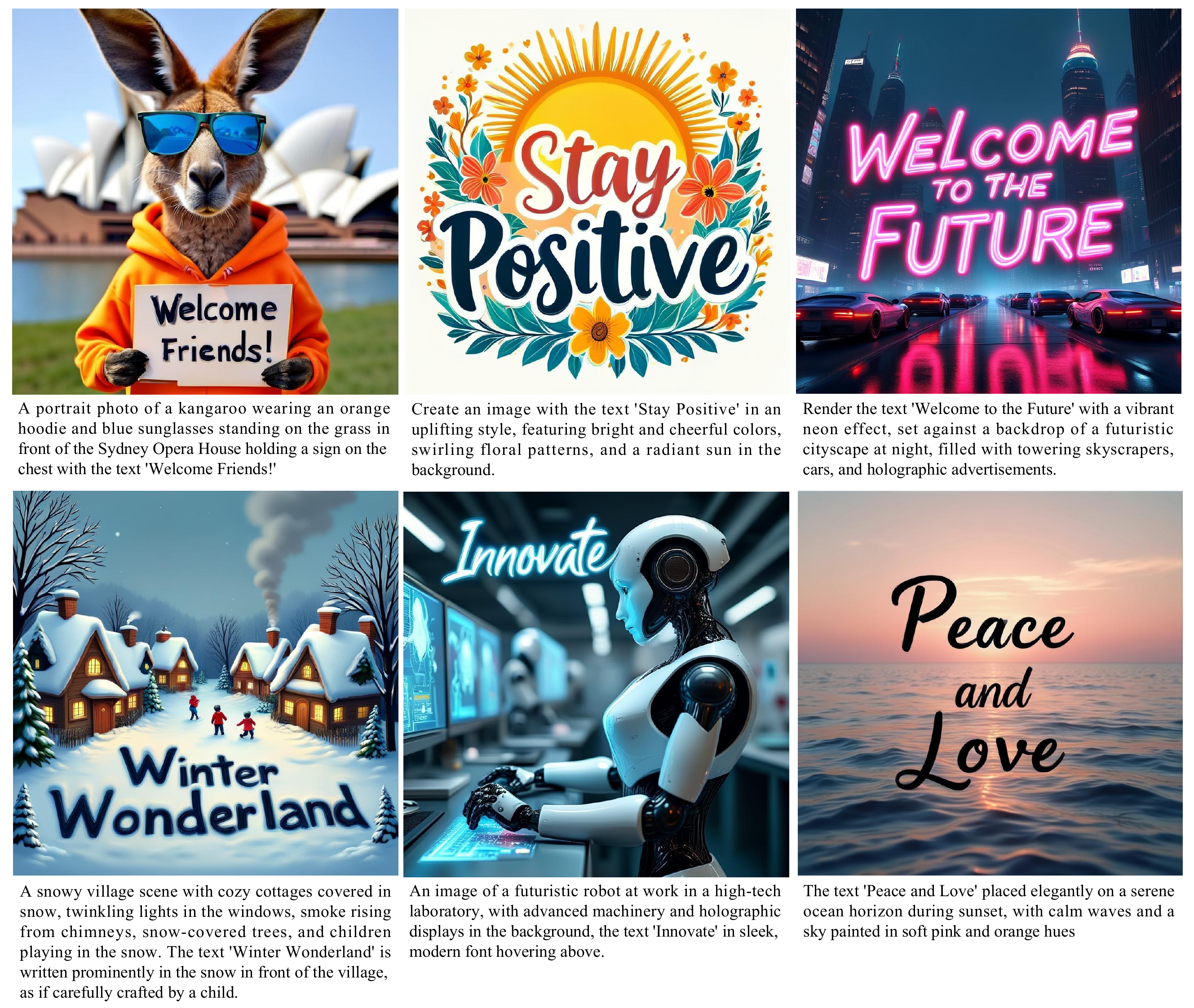}
   \vspace{-0.1cm}
   \caption{Text rendering results from our \methodNAME-2B model. \methodNAME-2B could generate text-consistent images following user prompts across diverse categories.}
   \label{fig:text_rendering}
\end{figure}
\vspace{-0.4cm}

\noindent\textbf{Inference Latency.}  
As shown in Tab. \ref{tab:hpsTable}, Infinity demonstrates a significant advantage in generation speed compared to diffusion models at around 2 billion parameters. Furthermore, our tests reveal that the speed advantage of \methodNAME becomes more substantial as the model size increases. \methodNAME achieves 7× faster inference latency compared to SD3.5 \cite{stable-diffusion3} at the same 8 billion parameters.

\vspace{-0.4cm}
\begin{table}[H]
\tiny
\centering
\captionsetup{skip=5pt} 
\caption{ \textbf{Human Preference Metrics and Inference Latency}. We compared our method with SoTA open-source models. \methodNAME achieved the best human preference results with the fastest speed.}
\label{tab:hpsTable}
\setlength{\tabcolsep}{2pt}
\resizebox{0.8\linewidth}{!}{
\begin{tabular}{lccccccc}
\toprule
\multirow{2}{*}{Methods} & \multirow{2}{*}{\# Params} & \multicolumn{2}{c}{ImageReward$\uparrow$} & \multicolumn{2}{c}{HPSv2.1$\uparrow$} & \multicolumn{2}{c}{Latency$\downarrow$} \\ \cmidrule(l){3-8}
 & & Rank & Score & Rank & Score & Rank & Time \\ \midrule
SD-XL~\cite{sdxl} & 2.6B & 4 & 0.600 & 4 & 30.06 & 4 & 2.7s \\
SD3-Medium~\cite{stable-diffusion3} & 2B & 3 & 0.871 & 3 & 30.91 & 3 & 2.1s \\
PixArt Sigma~\cite{chen2024pixart_sigma} & 630M & 2 & 0.872 & 2 & 31.47 & 2 & 1.1s \\
\cellcolor{mycolor_green}{\textbf{\methodNAME}} & \cellcolor{mycolor_green}{2B} & \cellcolor{mycolor_green}{1} & \cellcolor{mycolor_green}{\textbf{0.962}} & \cellcolor{mycolor_green}{1} & \cellcolor{mycolor_green}{\textbf{32.25}} & \cellcolor{mycolor_green}{1} & \cellcolor{mycolor_green}{\textbf{0.8s}} \\ 
\bottomrule
\end{tabular}
}
\end{table}

\vspace{-0.7cm}
\begin{table}[H]
\centering
\captionsetup{skip=5pt} 
\caption{ Comparison of memory consumption (GB) between different quantizers during training. As codebook dimension $d$ increases, MSR-BSQ shows significant advantages over MSR-LFQ, enabling nearly infinite vocabulary size of $2^{64}$.
}
\label{tab:memory-comparison}
\resizebox{0.63\linewidth}{!}{
\begin{tabular}{cccccc}
\toprule
Quantizer & $d=16$ & $d=18$ & $d=20$ & $d=32$ & $d=64$ \\
\midrule
LFQ & 37.6 & 53.7 & OOM & OOM & OOM \\
\textbf{BSQ} & 32.4 & 32.4 & 32.4 & 32.4 & 32.4 \\
\bottomrule
\end{tabular}
}
\end{table}

\vspace{-0.7cm}
\begin{table}[H]
\centering
\captionsetup{skip=5pt} 
\caption{ By scaling up visual tokenizer's vocabulary, discrete tokenizer surpasses continuous VAE of SD~\cite{ldm} on ImageNet-rFID.
}
\label{tab:scaling_vocabulary}
\setlength{\tabcolsep}{2pt}
\resizebox{.6\linewidth}{!}{
\begin{tabular}{lcccc}
\toprule
VAE (stride=16) & TYPE & IN-256 rFID$\downarrow$ & IN-512 rFID$\downarrow$ \\ \midrule
$V_{d}=2^{16}$ & Discrete & 1.22 & 0.31 \\
$V_{d}=2^{24}$ & Discrete & 0.75 & 0.30 \\
$V_{d}=2^{32}$ & Discrete & 0.61 & 0.23 \\
$V_{d}=2^{64}$ & Discrete &\textbf{0.33} & \textbf{0.15} \\
\midrule
SD VAE \cite{rombach2022high} & Contiguous & 0.87 & N/A \\
\bottomrule
\end{tabular}
}
\end{table}

\vspace{-0.7cm}
\begin{table}[H]
\centering
\captionsetup{skip=5pt} 
\caption{ IVC saves 99.95\% params and gets better performance to conventional classifier ($V_d=2^{16})$
}
\label{tab:ivc_classifier}
\resizebox{0.95\linewidth}{!}{
\begin{tabular}{lcccccc}
\toprule
Classifier & \# Params & vRAM & Recons. Loss$\downarrow$ & FID$\downarrow$ & ImageReward$\uparrow$ & HPSv2.1$\uparrow$ \\
\midrule
Convention & 124M & 2GB & 0.184 & 4.49 & 0.79 & 31.95 \\
\cellcolor{mycolor_green}{IVC} & \cellcolor{mycolor_green}{0.65M} & \cellcolor{mycolor_green}{10MB} & \cellcolor{mycolor_green}{0.180} & \cellcolor{mycolor_green}{3.83} & \cellcolor{mycolor_green}{0.91} & \cellcolor{mycolor_green}{32.31} \\
\bottomrule
\end{tabular}
}
\end{table}

\vspace{-0.7cm}
\begin{table}[H]
\centering
\captionsetup{skip=5pt} 
\caption{ Model architectures for scaling visual autoregressive modeling. Note that GFLOPs are rough values since they are affected by the length of the text prompt.
}
\label{tab:model_architecture}
\resizebox{0.6\linewidth}{!}{
\begin{tabular}{lcccc}
\toprule
\# Params & GFLOPs & Hidden Dimension & Heads & Layers \\
\midrule
125M & 30 & 768 & 8 & 12 \\
361M & 440 & 1152 & 12 & 16 \\
940M & 780 & 1536 & 16 & 24 \\
2.2B & 1500 & 2080 & 20 & 32 \\
4.7B & 2600 & 2688 & 24 & 40 \\
\bottomrule
\end{tabular}
}
\end{table}
\vspace{-0.6cm}

\subsection{Scaling Visual Tokenizer's Vocabulary}

\textbf{Scaling Up the Vocabulary Benefits Reconstruction.} Restricted by the vocabulary size, discrete VQ-VAEs have always lagged behind continuous ones, hindering the performance of AR-based T2I models. In this work, we successfully train a discrete VQ-VAE matching its continuous counterparts by scaling up the vocabulary size. As in Tab. \ref{tab:scaling_vocabulary}, we observe consistent rFID improvements as scaling up the vocabulary size from $2^{16}$ to $2^{64}$. It's noteworthy that our discrete tokenizer achieves a rFID of 0.61 on ImageNet 256$\times$256 when $V_{d}=2^{32}$, outperforming the continuous VAE of SD \cite{rombach2022high}.

\begin{figure}[h]
  \centering
   \includegraphics[width=1.0\linewidth]{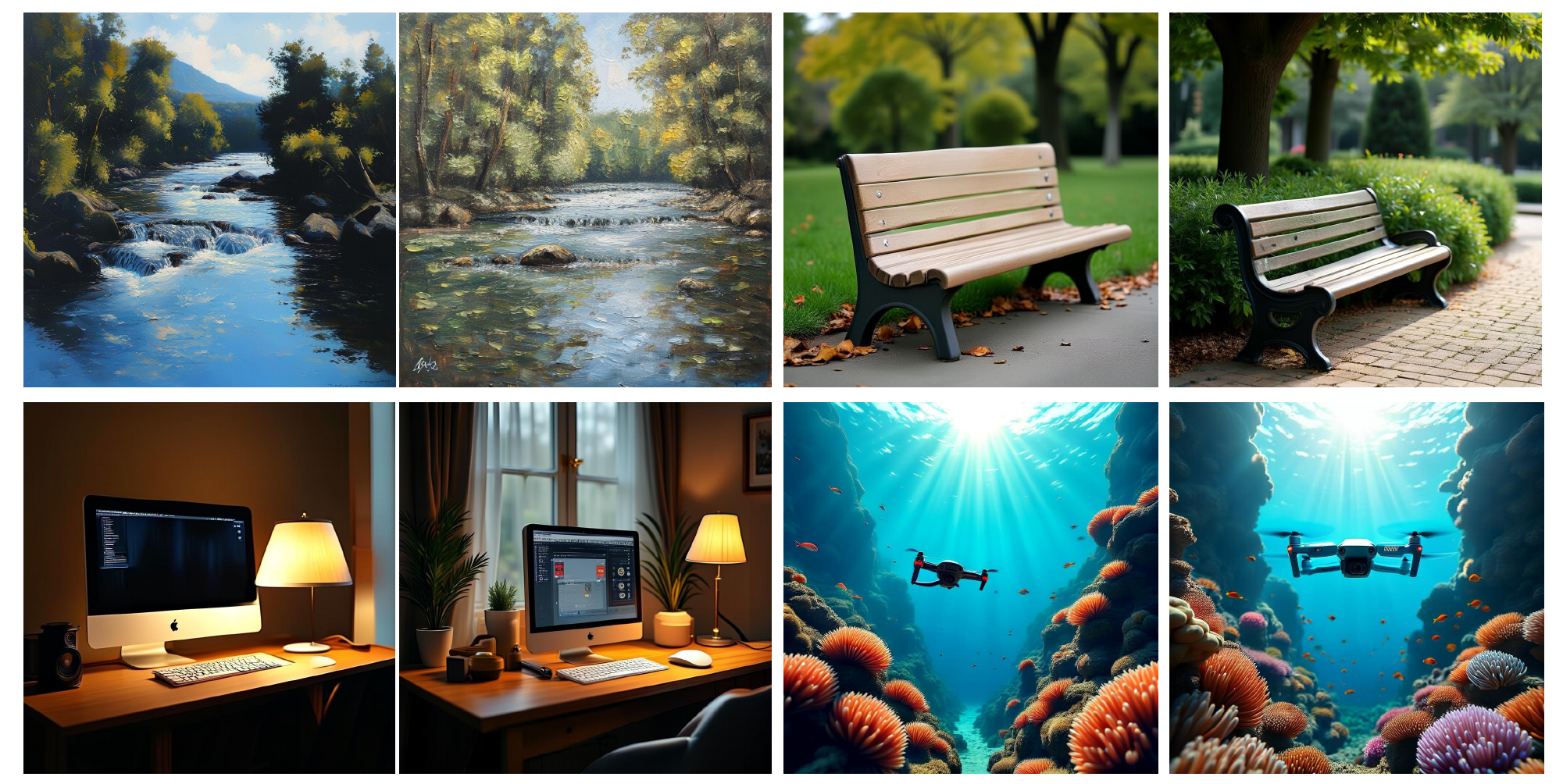}
   \caption{\textbf{Impact of Infinite-Vocabulary Classifier}. Predicting bitwise labels with the Infinite-Vocabulary Classifier (Right) generates images with richer details compared to predicting index-wise labels using a conventional classifier (Left).}
   \label{fig:ivc_vis}
\end{figure}

\begin{figure}[H]
  \centering
   \includegraphics[width=1.0\linewidth]{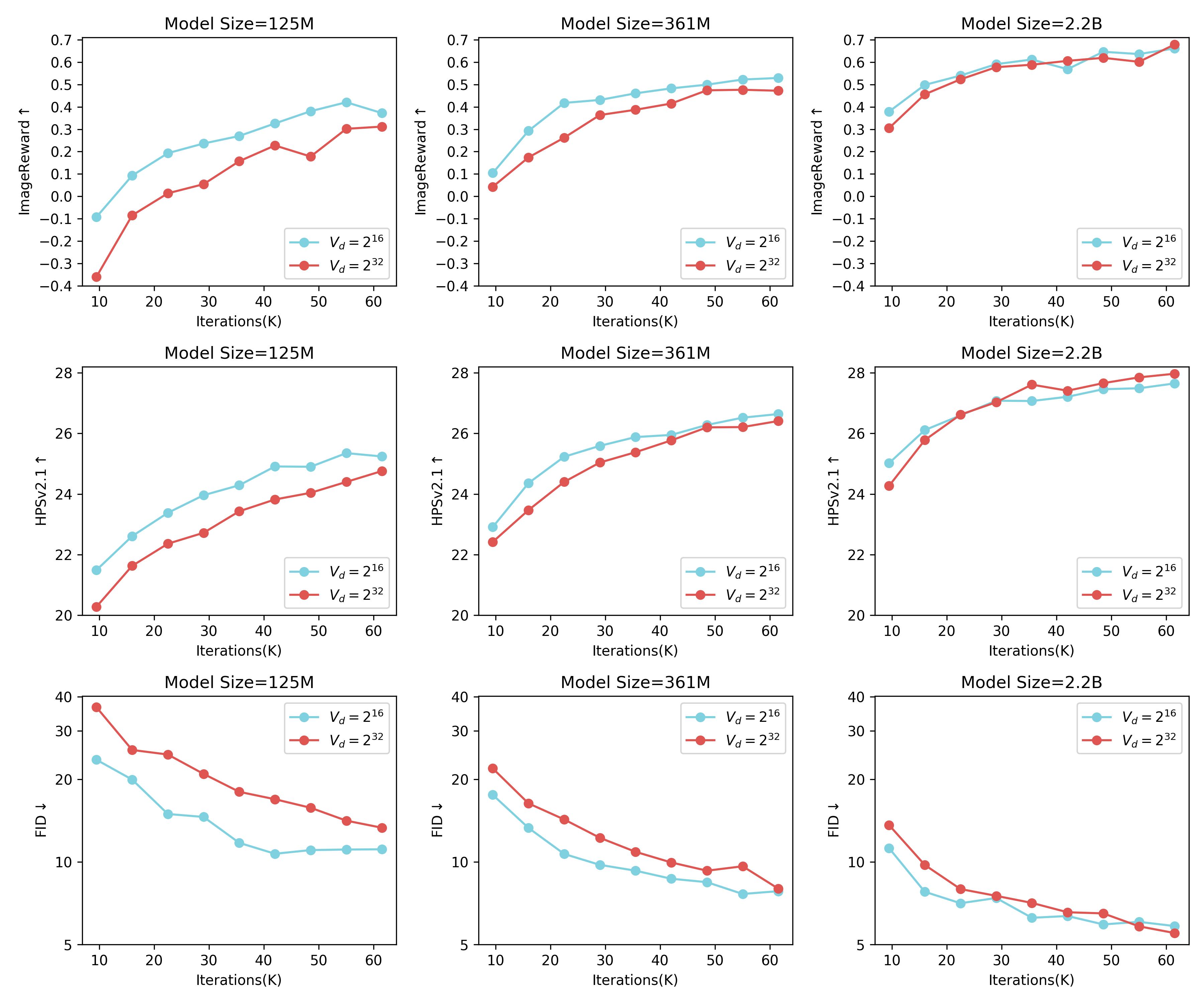}
   \vspace{-0.2cm}
   \caption{\textbf{Effects of Scaling Up the Vocabulary.} We analyze the impact of scaling the vocabulary size under consistent training hyperparameters throughout. Vocabulary size $V_d=2^{16}$ converges faster and achieves better results for small models (125M and 361M parameters). As we scale up the model size to 2.2B, \methodNAME with a vocabulary size $V_d=2^{32}$ beats that one with $V_d=2^{16}$. Experiment with 5M high-quality image-text pair data under $256\times 256$ resolution.}
   \label{fig:scaling_vocabulary}
\end{figure}

\noindent\textbf{Infinite Vocabulary Classifier Benefits Generation.} We compare predicting bit labels with IVC to predicting index labels using a conventional classifier under the vocabulary size of $2^{16}$, since a larger vocabulary causes OOM for the conventional classifier. We use the reconstruction loss on $\bm{R}_{k}$, FID on the validation set 
 and ImageReward for comprehensive evaluation. As shown in Tab.\ref{tab:ivc_classifier}, IVC achieves lower reconstruction loss and FID, suggesting IVC has better fitting capabilities. Beyond the quantitative results, training \methodNAME with IVC yields images with richer details as in Fig.\ref{fig:ivc_vis}, which is consistent with a higher ImageReward.

\begin{figure}[t]
  \centering
   \includegraphics[width=1.0\linewidth]{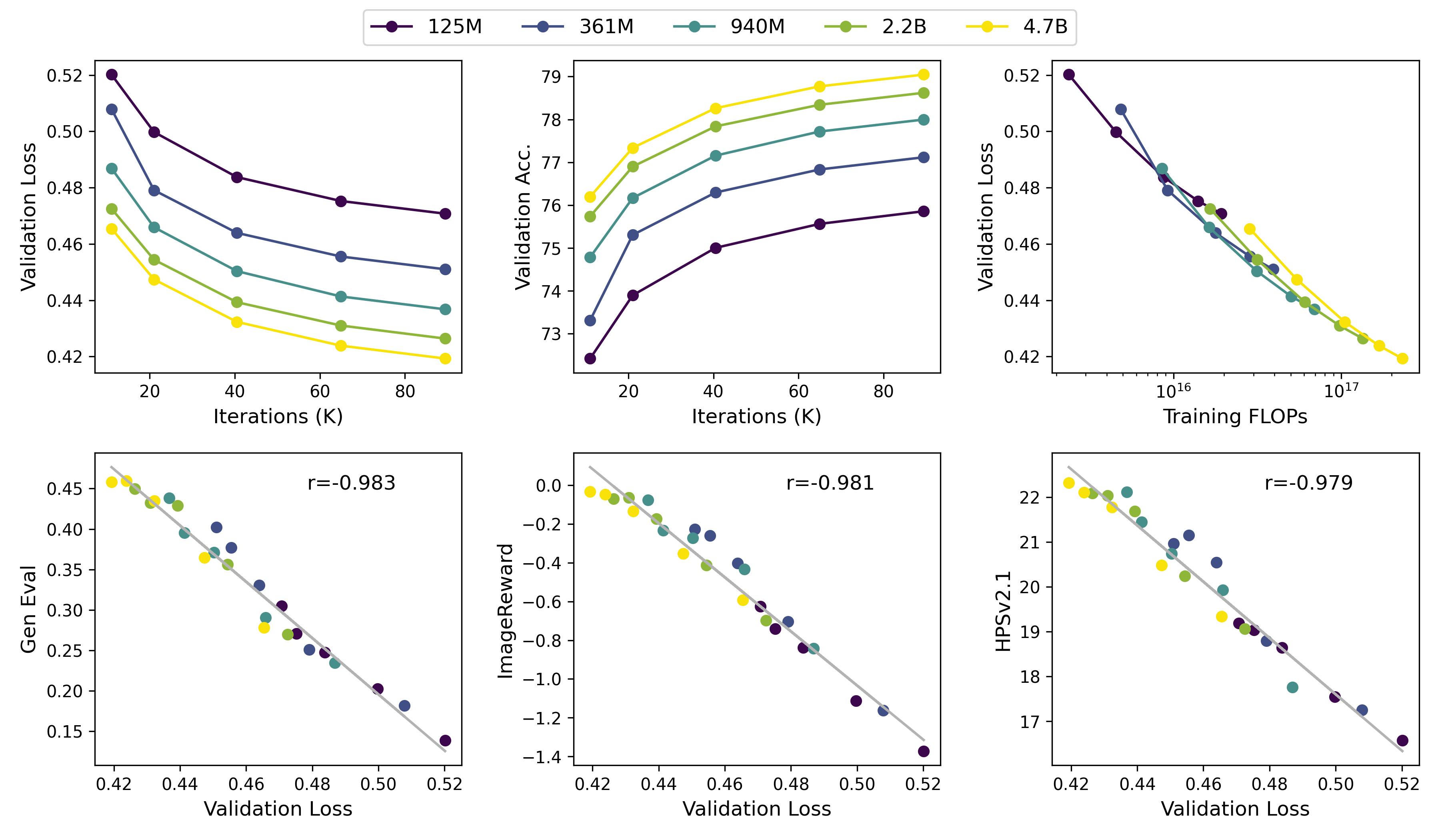}
   \caption{\textbf{Effects of Scaling Visual AutoRegressive Modeling.} We analyze the impact of scaling model size under consistent training hyperparameters throughout (Experiment with 10M pre-training data and $256\times 256$ resolution). Validation loss smoothly decreases as a function of the model size and training iterations. Besides, Validation loss is a strong predictor of overall model performance. There is a strong correlation between validation loss and holistic image evaluation metrics.}
   \label{fig:scaling_models}
\end{figure}

\subsection{Scaling Bitwise AutoRegressive Modeling}
\noindent\textbf{Scaling Up the Vocabulary Benefits Generation.} We then scale up the vocabulary size to $2^{32}$ during training the T2I model, which exceeds the range of the Int32 data type and can be considered infinitely large. In Fig.\ref{fig:scaling_vocabulary}, we illustrate the effect of scaling up the vocabulary from $2^{16}$ to $2^{32}$ for image generation. For small models (125M and 361M), the vocabulary size of $2^{16}$ converges faster and achieves better results. However, as we scaled up the transformer to 2.2B, the vocabulary size of $2^{32}$ beats $2^{16}$ after 40K iterations. Therefore, it's worthwhile to scale up the vocabulary along with scaling up the transformer. As illustrated in Tab.\ref{tab:genEvalSotaTable},\ref{tab:hpsTable}, with infinite vocabulary and IVC, \methodNAME achieves superior performance among various benchmarks, elevating the ceiling of AR visual generation. 

\begin{figure}[!h]
  \centering
   \includegraphics[width=1.0\linewidth]{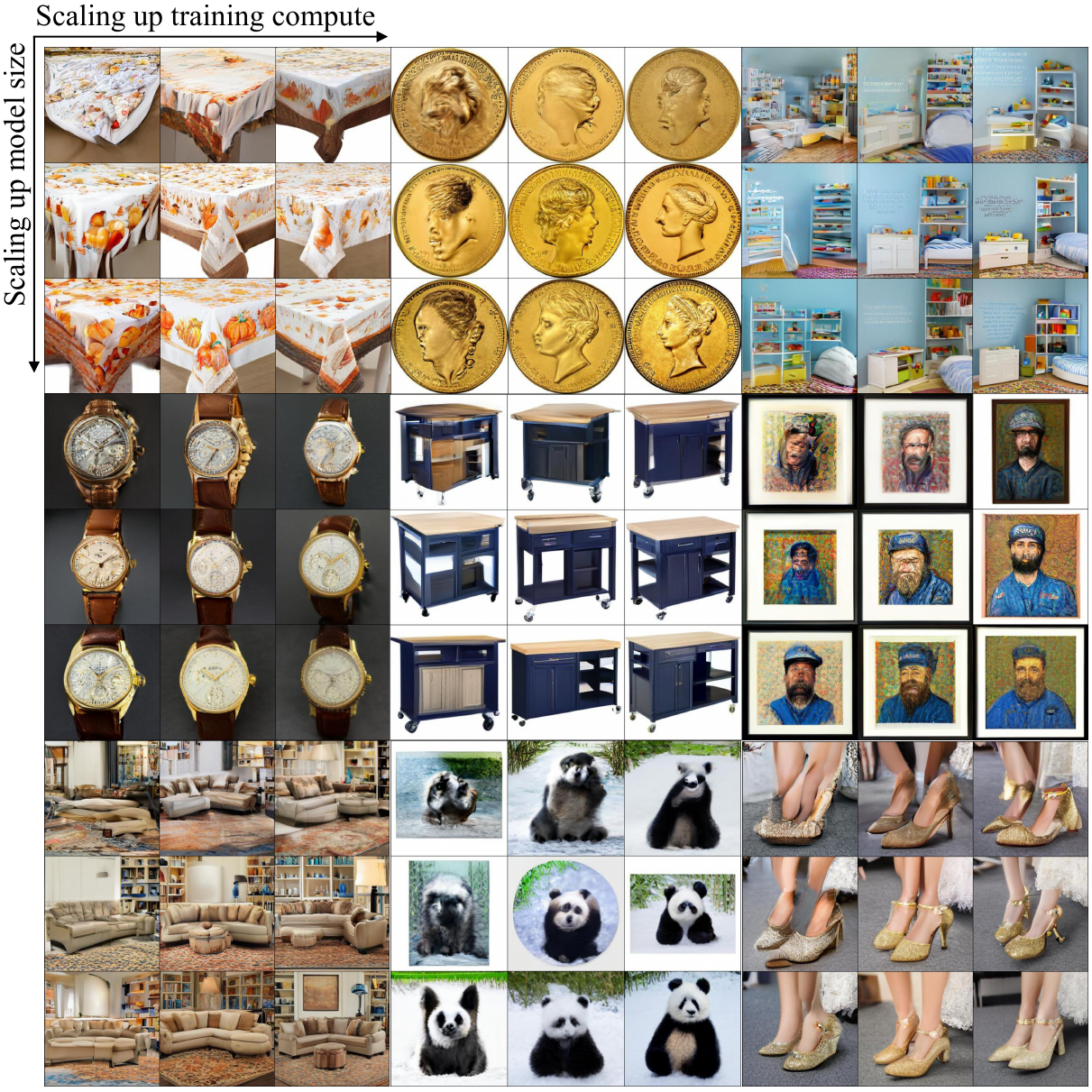}
   \caption{Semantics and visual quality improve consistently with scaling up model size and training compute. Zoom in for better comparison.}
   \label{fig:scaling_models_images}
\end{figure}

\noindent\textbf{Scaling Up Transformer Benefits Generation.} In Fig.\ref{fig:scaling_models}, we depict the validation loss against the total training iterations and computational FLOPs for various model sizes of \methodNAME. The detailed model architectures for different sizes can be found in Tab.\ref{tab:model_architecture}. We consistently notice a reduction in validation loss with an increase in training steps and computational FLOPs. Nevertheless, the advantages gained from training smaller models for extended periods lag behind those obtained from training larger models for shorter durations. This trend aligns with findings in language models, emphasizing the promising outlook for increasing model sizes with appropriate training.

In Fig.\ref{fig:scaling_models}, we plot GenEval, ImageReward, and HPSv2 scores against validation loss for different model sizes ranging from 125M to 4.7B. We observe a strong correlation between validation loss and evaluation metrics. To further quantify their correlation, we calculate the Pearson correlation coefficients through linear regression. The correlation coefficients for GenEval, ImageReward, and HPSv2 are -0.983, -0.981, and -0.979, respectively. These results demonstrate a nearly linear correlation between validation loss and the evaluation metrics when scaling up model sizes from 125M to 4.7B. This promising phenomenon encourages us to scale up \methodNAME to achieve better performance.

\noindent\textbf{Visualization of Scaling Effects.} To delve deeper into the scaling effect of \methodNAME, we compare a set of generated 256$\times$256 images of three model sizes (125M, 940M, 4.7B) across three distinct training schedules (10K, 40K, 90K iterations) as illustrated in Fig.\ref{fig:scaling_models_images}. The semantics and visual quality of generated images improve steadily when scaling up model size and training compute, which is consistent with the scaling behaviors of \methodNAME.

\begin{figure}[h]
  \centering
   \includegraphics[width=1.0\linewidth]{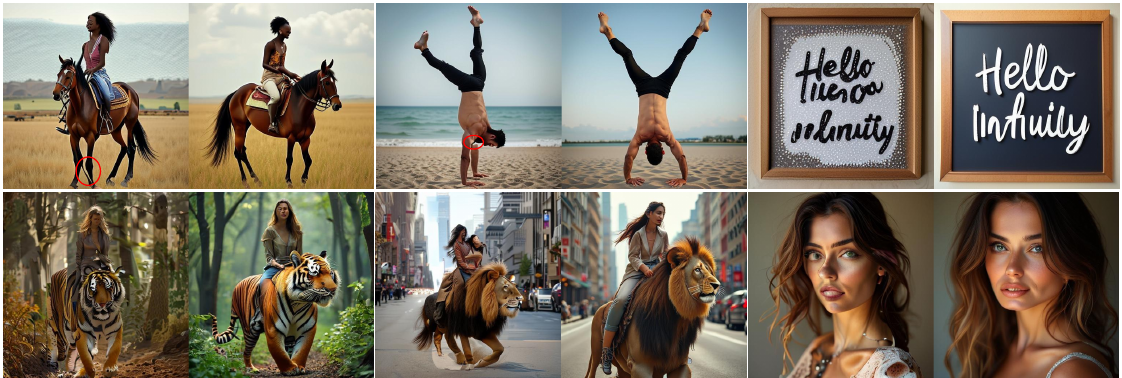}
   \caption{\textbf{Impact of Self-Correction}. Teacher-forcing training introduces great train-test discrepancy which degrades performance during inference (left). Bitwise Self-Correction auto-corrects mistakes and thus generates better results (right). Decoding with $\tau=1$ and $cfg=3$.}
   \label{fig:self_correction}
\end{figure}

\subsection{Bitwise Self-Correction}
In Tab.\ref{tab:self_correction} and Fig.\ref{fig:self_correction}, we list the evaluation metrics and present images generated by models trained using teacher-forcing and bitwise self-correction methods. Substantial advantages are observed after applying bitwise self-correction. Furthermore, we prove that the significant advantages are primarily driven by the self-correction mechanism rather than applying flipping. As shown in Tab.\ref{tab:self_correction}, simply random flipping $\bm{R}_k$ doesn't bring improvements. Self-Correction imitates prediction errors and applies re-quantification to correct them. We emphasize that Self-Correction is essential for AR-based T2I models since it empowers models to correct errors automatically, significantly mitigating the train-test discrepancy.

\vspace{-0.7cm}
\begin{table}[h]
\centering
\captionsetup{skip=5pt} 
\caption{ Bitwise Self-Correction makes significant improvements. Experiment with 5M high-quality data and $512\times 512$ resolution. FID is measured on the validation set with 40K images. Decoding with $\tau=1$ and $cfg=3$.
}
\label{tab:self_correction}
\resizebox{0.7\linewidth}{!}{
\begin{tabular}{lcccc}
\toprule
Method & FID$\downarrow$ & ImageReward$\uparrow$ & HPSv2.1$\uparrow$ \\
\midrule
Baseline & 9.76 & 0.52 & 29.53 \\
Baseline + Random Flip & 9.69 & 0.52 & 29.20 \\
\cellcolor{mycolor_green}{Baseline + Bitwise Self-Correction} & \cellcolor{mycolor_green}{3.48} & \cellcolor{mycolor_green}{0.76} & \cellcolor{mycolor_green}{30.71} \\
\bottomrule
\end{tabular}
}
\end{table}

\subsection{Ablation Studies}

\textbf{Optimal Strength for Bitwise Self-Correction.}
Bitwise Self-Correction mitigates the train-test discrepancy caused by teacher-forcing training. Here we delve into the optimal strength for applying bitwise self-correction in Tab.\ref{tab:self_correction_strength}. We empirically find that mistake imitation that is too weak (10\% and 20\%) fails to fully leverage the potential of Bitwise Self-Correction. Random flipping 30\% bits yields the best results.

\vspace{-0.3cm}
\begin{table}[H]
\centering
\captionsetup{skip=5pt} 
\caption{ Comparison between different strengths of Bitwise Self-Correction. Experiment with 5M high-quality data and $512\times 512$ resolution. Decoding with $\tau=1$ and $cfg=3$.
}
\label{tab:self_correction_strength}
\resizebox{0.7\linewidth}{!}{
\begin{tabular}{lcccc}
\toprule
Method & FID$\downarrow$ & ImageReward$\uparrow$ & HPSv2.1$\uparrow$ \\
\midrule
w/o Bitwise Self-Correction & 9.76 & 0.515 & 29.53 \\
\midrule
Bitwise Self-Correction ($p=10\%$) & 3.45 & 0.751 & 30.47 \\
Bitwise Self-Correction ($p=20\%$) & 3.48 & 0.763 & 30.71 \\
Bitwise Self-Correction ($p=30\%$) & \textbf{3.33} & \textbf{0.775} & \textbf{31.05} \\
\bottomrule
\end{tabular}
}
\end{table}
\vspace{-0.4cm}

\begin{figure}[H]
  \centering
   \includegraphics[width=1.0\linewidth]{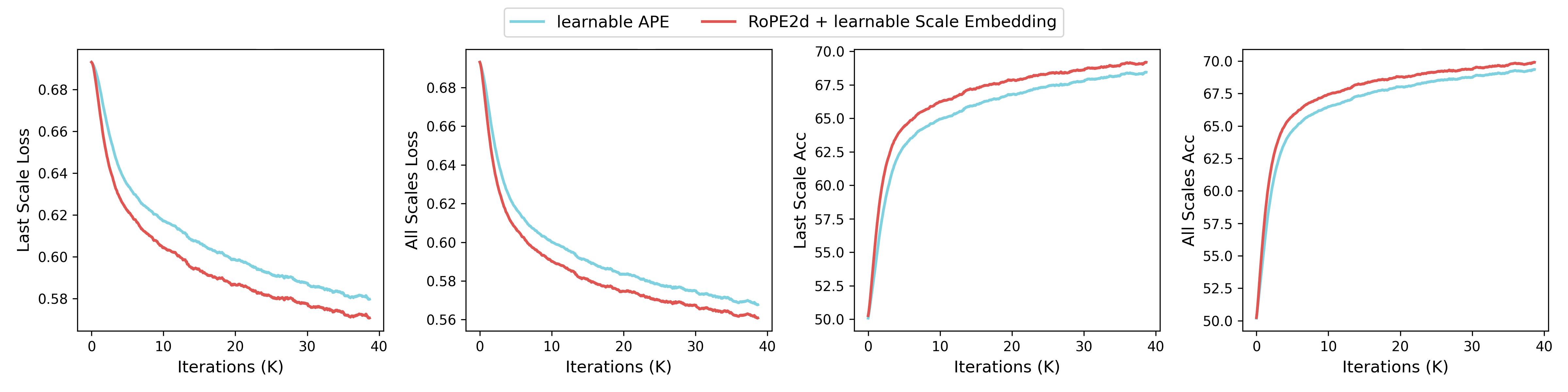}
   \vspace{-0.6cm}
   \caption{Comparison between learnable APE and our positional embeddings. Our method, \emph{i.e.}, applying RoPE2d along with learnable scale embeddings on features of each scale, converges faster and reaches higher training accuracy.}
   \label{fig:pe_ablation}
\end{figure}

\noindent\textbf{Positional Embedding.}
Learnable APE adopted in VAR~\cite{VAR} brings too many parameters and gets confused when the sequence length varies. However, the sequence length changes frequently when training with various aspect ratios. Simply applying RoPE2d~\cite{ROPE2d} or normalized RoPE2d~\cite{STAR} can not distinguish features from different resolutions. In this work, we apply RoPE2d and learnable scale embeddings on features of each scale. RoPE2d preserves the intrinsic 2D structure of images. Learnable scale embeddings avoid confusion between features of different scales. To verify the effectiveness, we compare it with the learnable APE in Fig.\ref{fig:pe_ablation}. It's obvious that applying RoPE2d along with learnable scale embeddings on features of each scale converges faster and reaches higher training accuracy. 

\begin{figure}[H]
  \centering
   \includegraphics[width=0.9\linewidth]{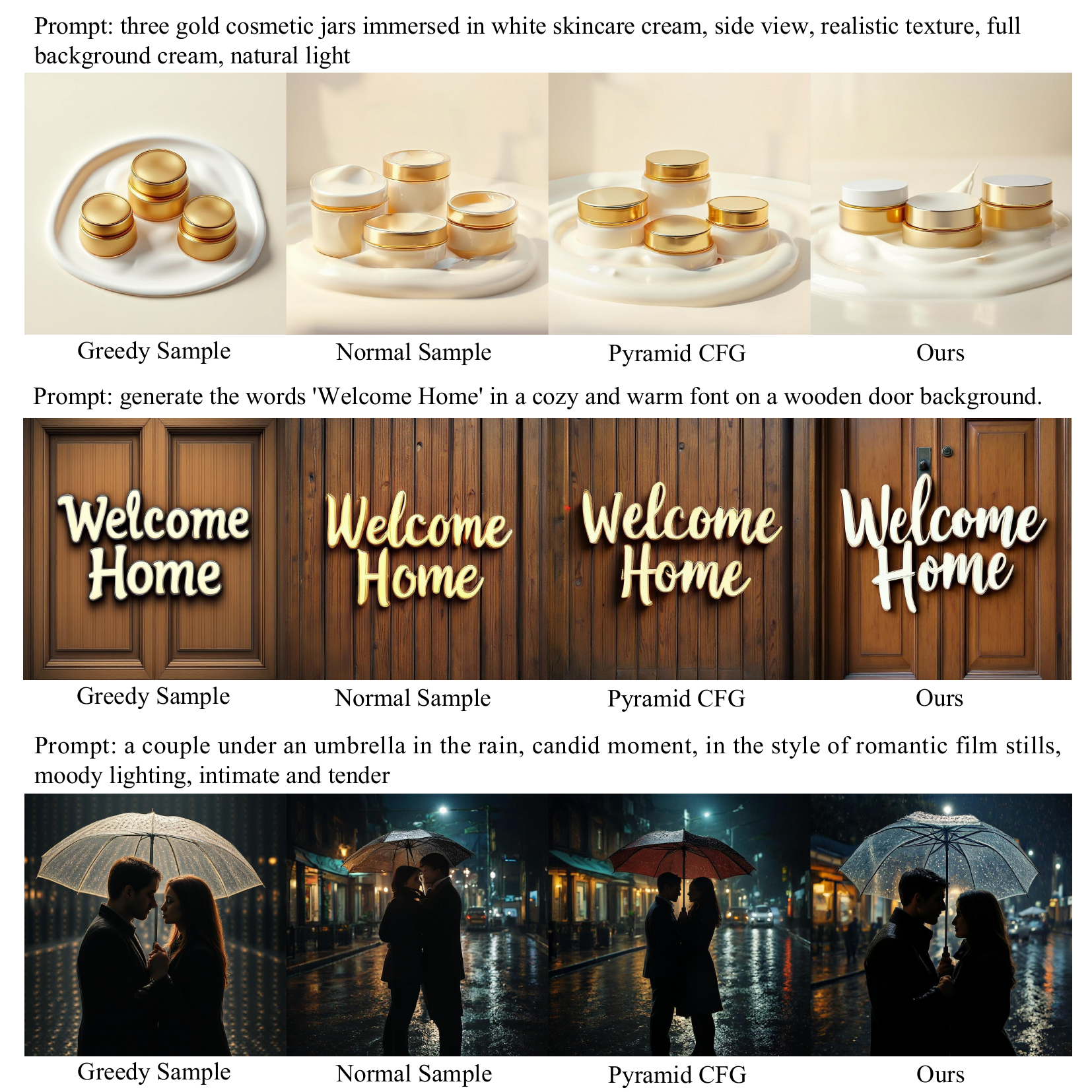}
   \vspace{-0.2cm}
   \caption{Comparison of different sampling methods. In contrast to Greedy Sample, Normal Sample and Pyramid Sample, our method could generate images with richer details and higher text-image alignments.}
   \label{fig:sample}
\end{figure}

\noindent\textbf{Decoding.} Decoding is crucial for improving generation quality. VAR adopts the pyramid Classifer-Free Guidance (CFG) on predicted logits. That is, the strength of CFG increases linearly as the scale goes from 1 to $K$. Such a pyramid scheme is used to tackle the issue of the model collapsing frequently when applying large CFG at early scales. We found that \methodNAME supports large CFG values even in very early scales equipped with Bitwise Self-Correction. Since \methodNAME is more robust to sampling, we revisit different decoding methods and find the best as illustrated in Tab.\ref{tab:decode}. We visualize the comparison results of different decoding methods in Fig.\ref{fig:sample}. We achieve the best generation results.

\vspace{-0.3cm}
\begin{table}[H]
\centering
\captionsetup{skip=5pt} 
\caption{ Comparison between different decoding methods.}
\label{tab:decode}
\resizebox{0.82\linewidth}{!}{
\begin{tabular}{llcccc}
\toprule
Method & Param & FID$\downarrow$ & ImageReward$\uparrow$ & HPSv2.1$\uparrow$ \\
\midrule
Greedy Sampling & $\tau=0.01, cfg=1$ &  9.97 & 0.397 & 30.98 \\
Normal Sampling & $\tau=1.00, cfg=1$ &  4.84 & 0.706 & 31.59 \\
Pyramid CFG & $\tau=1.00, cfg=1 \to 3 $ &  3.48 & 0.872 & \textbf{32.48} \\
Pyramid CFG & $\tau=1.00, cfg=1 \to 5 $ &  2.98 & 0.929 & 32.32 \\
CFG on features & $\tau=1.00, cfg=3$ & 3.00 & 0.953 & 32.13 \\
CFG on logits & $\tau=1.00, cfg=3$  & 2.91 & 0.952 & 32.31 \\
CFG on logits (Ours) & $\tau=1.00, cfg=4$  & \textbf{2.82} & \textbf{0.962} & 32.25 \\
\bottomrule
\end{tabular}
}
\end{table}
\vspace{-0.5cm}

\section{Conclusion}
\label{sec:conclusion}

We introduce \methodNAME, a bitwise visual autoregressive model to perform Text-to-Image generation. \methodNAME is a pioneering framework for bitwise token modeling with the IVC and self-correction innovation. Extensive qualitative and quantitative results demonstrate \methodNAME significantly raised the upper limit for Autogressive Text-To-Image generative models, matching or surpassing leading diffusion models. We believe our framework, \methodNAME, will substantially promote the development of autoregressive visual modeling and inspire the community for faster and more realistic generation models.

\section{Acknowledges}
Many colleagues from ByteDance supported this work. We are grateful to Guanyang Deng for his efforts in data processing. We also thank Chongxi Wang and Taekmin Kim for their contributions to model deployment. Special thanks to Xiaoxiao Qin for her work in human preference evaluation. Additionally, we are thankful to Hui Wu, Fu Li, Xing Wang, Hongxiang Hao, and Chuan Li for their contributions to infrastructure.


\clearpage
\newpage

\clearpage
{
\small
\bibliographystyle{plain}
\bibliography{infinity_main}
}

\newpage
\appendix
\clearpage
\setcounter{page}{1}

\section{Predefined Scale Schedules}
As listed in Tab.\ref{tab:patch_num_list}, for each aspect ratio $r$, we predefine a specific scale schedule $\{(h^r_1,w^r_1),...,(h^r_K,w^r_K)\}$. We ensure that the aspect ratio of each tuple $(h^r_k,w^r_k)$ is approximately equal to $r$, especially in the latter scales. Additionally, for different aspect ratios at the same scale $k$, we keep the area of $h^r_k \times w^r_k$ to be roughly equal, ensuring that the training sequence lengths are roughly the same. We adopt buckets to support training various aspect ratios at the same time. The consistent sequence lengths of different aspect ratios improve training efficiency. During the inference stage, \methodNAME could generate photo-realistic images covering common aspect ratios (1:1, 16:9, 4:3, \emph{etc.}) as well as special aspect ratios (1:3, 3:1, \emph{etc.}) following the predefined scale schedules.

\begin{table*}[h]
\tiny
\centering
\setlength{\tabcolsep}{2pt}
\captionsetup{skip=5pt} 
\caption{Predefined scale schedules $\{(h^r_1,w^r_1),...,(h^r_K,w^r_K)\}$ for different aspect ratios. Following the text guided next-scale prediction scheme, \methodNAME takes $K$=13 scales to generate a $1024\times 1024$ (or other aspect ratio) image.
}\label{tab:patch_num_list}
\resizebox{.99\linewidth}{!}{
\begin{tabular}{ccccccccccccccc}
\toprule
Aspect Ratio & Resolution & \multicolumn{13}{c}{Scale Schedule} \\
\midrule

1.000 (1:1) & 1024$\times$1024 & (1,1) & (2,2) & (4,4)&(6,6)&(8,8)&(12,12)&(16,16)&(20,20)&(24,24)&(32,32)&(40,40)&(48,48)&(64,64) \\

0.800 (4:5) & 896$\times$1120 & (1,1)&(2,2)&(3,3)&(4,5)&(8,10)&(12,15)&(16,20)&(20,25)&(24,30)&(28,35)&(36,45)&(44,55)&(56,70) \\

1.250 (5:4) & 1120$\times$896 & (1,1)&(2,2)&(3,3)&(5,4)&(10,8)&(15,12)&(20,16)&(25,20)&(30,24)&(35,28)&(45,36)&(55,44)&(70,56) \\

0.750 (3:4) & 864$\times$1152 & (1,1)&(2,2)&(3,4)&(6,8)&(9,12)&(12,16)&(15,20)&(18,24)&(21,28)&(27,36)&(36,48)&(45,60)&(54,72) \\

1.333 (4:3) & 1152$\times$864 & (1,1)&(2,2)&(4,3)&(8,6)&(12,9)&(16,12)&(20,15)&(24,18)&(28,21)&(36,27)&(48,36)&(60,45)&(72,54) \\

0.666 (2:3) & 832$\times$1248 & (1,1)&(2,2)&(2,3)&(4,6)&(6,9)&(10,15)&(14,21)&(18,27)&(22,33)&(26,39)&(32,48)&(42,63)&(52,78) \\

1.500 (3:2) & 1248$\times$832 & (1,1)&(2,2)&(3,2)&(6,4)&(9,6)&(15,10)&(21,14)&(27,18)&(33,22)&(39,26)&(48,32)&(63,42)&(78,52) \\

0.571 (4:7) & 768$\times$1344 & (1,1)&(2,2)&(3,3)&(4,7)&(6,11)&(8,14)&(12,21)&(16,28)&(20,35)&(24,42)&(32,56)&(40,70)&(48,84) \\

1.750 (7:4) & 1344$\times$768 & (1,1)&(2,2)&(3,3)&(7,4)&(11,6)&(14,8)&(21,12)&(28,16)&(35,20)&(42,24)&(56,32)&(70,40)&(84,48) \\

0.500 (1:2) & 720$\times$1440 & (1,1)&(2,2)&(2,4)&(3,6)&(5,10)&(8,16)&(11,22)&(15,30)&(19,38)&(23,46)&(30,60)&(37,74)&(45,90) \\

2.000 (2:1) & 1440$\times$720 & (1,1)&(2,2)&(4,2)&(6,3)&(10,5)&(16,8)&(22,11)&(30,15)&(38,19)&(46,23)&(60,30)&(74,37)&(90,45) \\

0.400 (2:5) & 640$\times$1600 & (1,1)&(2,2)&(2,5)&(4,10)&(6,15)&(8,20)&(10,25)&(12,30)&(16,40)&(20,50)&(26,65)&(32,80)&(40,100) \\

2.500 (5:2) & 1600$\times$640 & (1,1)&(2,2)&(5,2)&(10,4)&(15,6)&(20,8)&(25,10)&(30,12)&(40,16)&(50,20)&(65,26)&(80,32)&(100,40) \\

0.333 (1:3) & 592$\times$1776 & (1,1)&(2,2)&(2,6)&(3,9)&(5,15)&(7,21)&(9,27)&(12,36)&(15,45)&(18,54)&(24,72)&(30,90)&(37,111) \\

3.000 (3:1) & 1776$\times$592 & (1,1)&(2,2)&(6,2)&(9,3)&(15,5)&(21,7)&(27,9)&(36,12)&(45,15)&(54,18)&(72,24)&(90,30)&(111,37) \\
\bottomrule
\end{tabular}
}
\end{table*}

\begin{figure}[!h]
  \centering
   \includegraphics[width=1\linewidth]{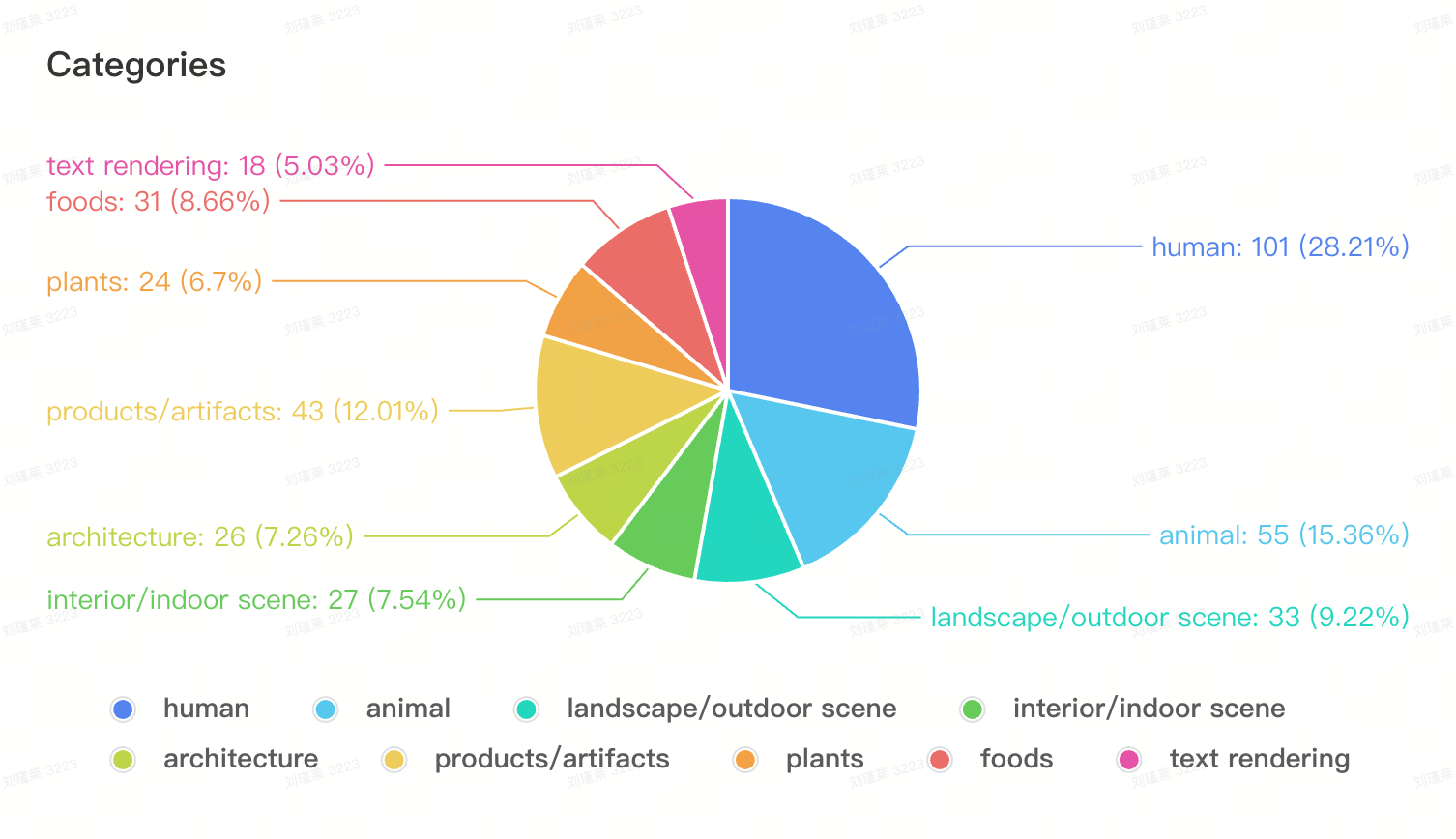}
    \vspace{-3mm}
   \caption{Distribution of Prompt Categories}
   \vspace{-3mm}
   \label{fig:categories}
\end{figure}

\begin{figure}[!h]
  \centering
   \includegraphics[width=1\linewidth]{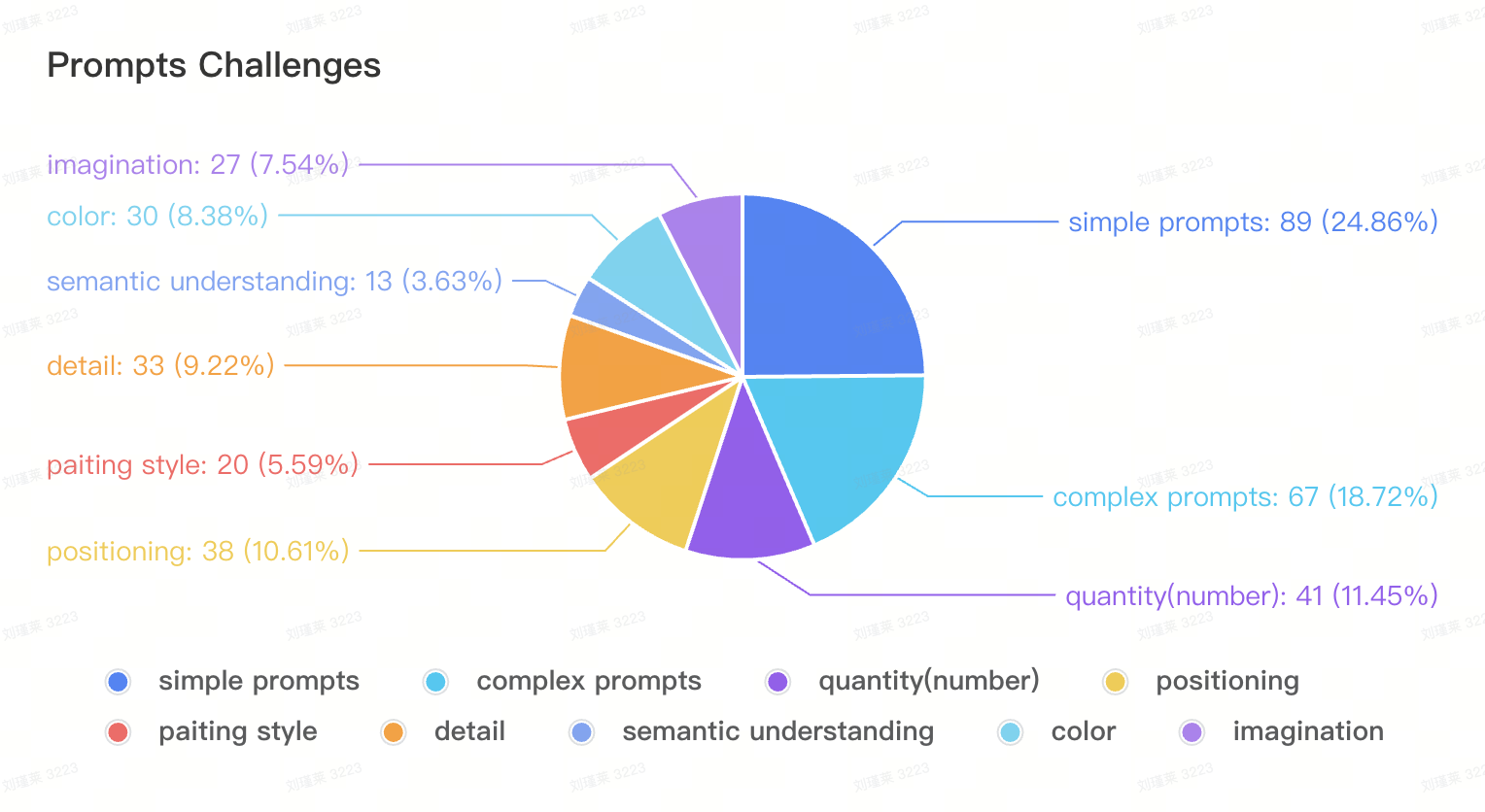}
    \vspace{-3mm}
   \caption{Distribution of Prompts Challenges}
   \vspace{-3mm}
   \label{fig:prompt_challenges}
\end{figure}

\section{Human Preference Evaluation}
In order to measure the overall performance, we have conducted a human preference evaluation. We build a website and recruit volunteers to rank the generated images from different T2I models.

\textbf{Prompts.} We have collected 360 prompts in total, including prompts randomly sampled from Parti \cite{parti} and other human-written prompts. As illustrated in Fig.\ref{fig:categories}, these prompts are divided into nine categories, such as human (28\%), animal (15\%), products/artifacts (12\%), landscape (9\%), foods, indoor scene, architecture, plants, and text rendering. It is worth noting that we incorporate a variety of human-related prompts, such as faces, bodies, and movements, in the human category as a supplement to the Parti prompts. In Fig.\ref{fig:prompt_challenges}, we also list the challenges of these prompts, which includes simple prompts, complex prompts, quantity, positioning \& perspective, painting style, detail, semantic understanding, color, and imagination. These statistics demonstrate that the prompts used for evaluation are balanced, covering various categories and challenges well.

\textbf{Generated Images.} We compare \methodNAME with four open-source models: PixArt-Sigma~\cite{chen2024pixart_sigma}, SD3-Medium~\cite{stable-diffusion3}, SDXL~\cite{sdxl}, and HART~\cite{tang2024hart}. The images of other models are generated by running their official inference code. No cherry-picking for any models.

\textbf{Human Evaluation.} For the human evaluation process, we build a website which presents two images from two anonymous models at the same time. There is one image generated by \methodNAME while the other is from other four models. Volunteers are required to pick a better one from two images in terms of \emph{overall quality, prompt following, and visual aesthetics}, respectively. Besides the aforementioned criterion, we make sure each side-by-side comparison is evaluated by at least two volunteers to reduce human bias. We filter out pairs with opposite results evaluated by two volunteers. These contradictory pairs are sent to a third volunteer to assess. Then we take the consensus from three as the final results. Note that the whole process of human evaluation is completely double-blind. That is, a volunteer doesn't know which model it is, as well as other volunteers' results when performing a side-by-side comparison.

\textbf{Results.} As in Fig.6 of the submitted manuscript, we observe a remarkable human preference for \methodNAME over the other four open-source models. Especially for the comparison with HART~\cite{tang2024hart} (another SOTA AR-based model), \methodNAME earns 90.0\%, 83.9\%, and 93.2\% win rate in terms of overall quality, prompt following, and visual aesthetics, respectively. As for the diffusion family, \methodNAME earns 76.0\%, 79.0\%, 66.0\% win rate to PixArt-Sigma, SDXL and SD3-Medium, respectively. What's more, \methodNAME reaches 71.1\% win rate towards SD3-Medium regarding visual aesthetics. These results reveal that \methodNAME is more capable of generating visually appealing images. We attribute these great advantages to the proposed bitwise modeling, which has lifted the upper limits of AR models by large margins.

\begin{figure*}[h]
  \centering
   \includegraphics[width=1.0\linewidth]{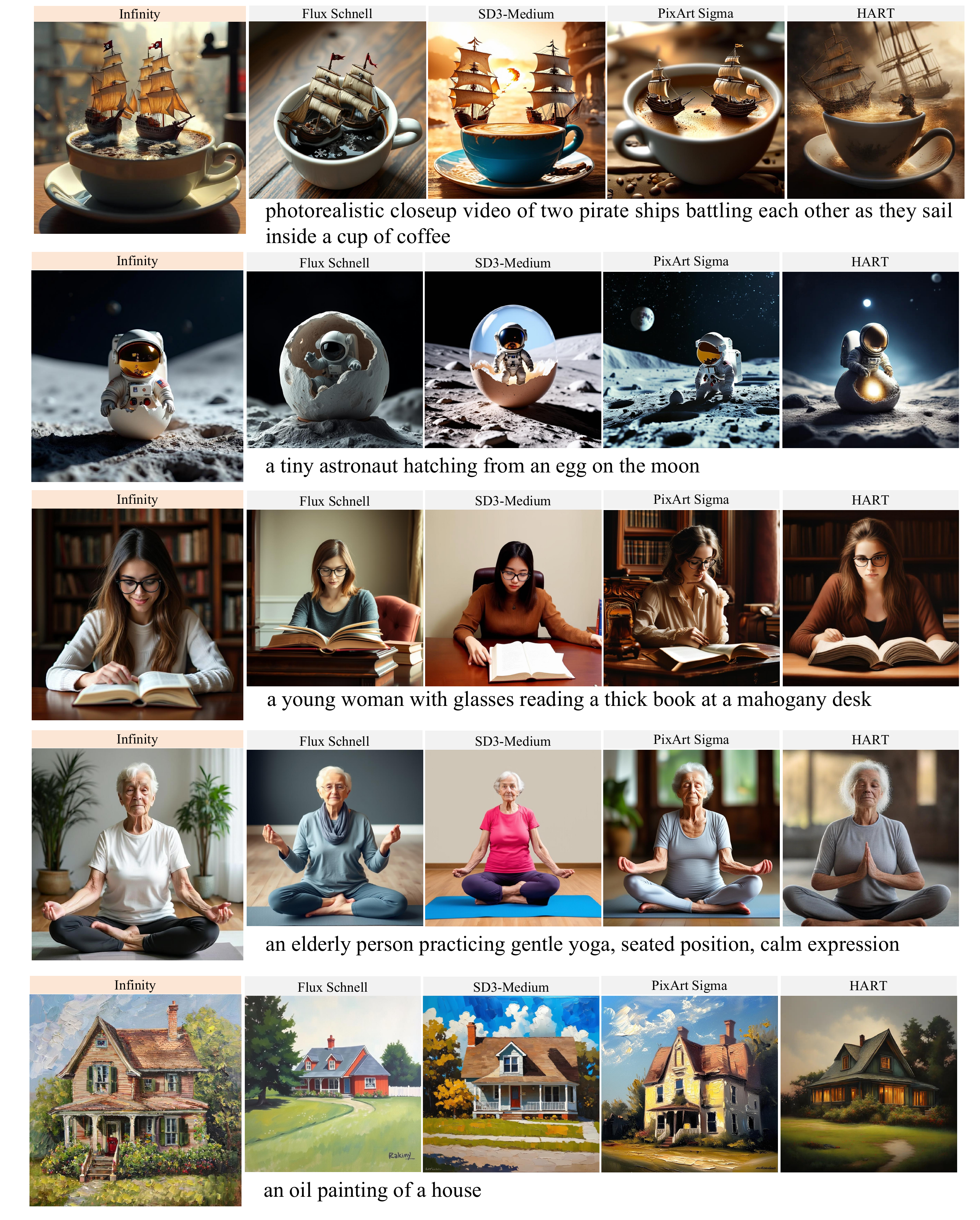}
   \caption{T2I qualitative comparison among our \methodNAME-2B model and the other four open-source models. Here we select three diffusion models (Flux Schnell, SD3-Medium and PixArt Sigma), one AR model (HART) for comparison. Zoom in for better comparsion.}
   \label{fig:qualitative_comparison}
\end{figure*}

\section{More Qualitative Results}
Fig.\ref{fig:qualitative_comparison} shows the qualitative comparison results among \methodNAME and other top-tier models. The images of other models are obtained either by querying their open-source demo website (HART \cite{tang2024hart}) or running their official inference code locally (Flux-Schnell~\cite{FLUX}, SD3-Medium \cite{stable-diffusion3}, and PixArt Sigma \cite{chen2024pixart_sigma}). Whether a thumbnail or a zoom-in image, we observe significant differences among the generated images from different models. In particular, the AR model like HART generates images with fewer details, blurred human faces and texture-less background compared to diffusion models. In contrast, \methodNAME overcomes those shortcomings of AR models and generates comparable or better images when compared to diffusion models like Flux-Schnell, SD3-Medium, and PixArt Sigma. For the first and second examples, \methodNAME adheres to the text prompts better than SD3-Medium, HART, and PixArt-Sigma. For the third and fourth examples, \methodNAME performs better in human hands and legs. For the last example, \methodNAME and PixArt Sigma have successfully generated images in an oil painting style while the other three failed. Flux Schnell performs worst in this example.

\end{document}